\documentclass[10pt,twocolumn,letterpaper]{article}

\PassOptionsToPackage{dvipsnames}{xcolor}
\usepackage[accsupp]{axessibility}
\usepackage{multirow}
\usepackage{graphicx}
\usepackage{pifont}
\usepackage{amssymb}
\usepackage{algorithm}
\usepackage[pagenumbers]{cvpr} % To force page numbers, e.g. for an arXiv version

\definecolor{cvprblue}{rgb}{0.21,0.49,0.74}
\usepackage[pagebackref,breaklinks,colorlinks,allcolors=cvprblue]{hyperref}
\usepackage{tikz}
\newcommand*\circled[1]{\tikz[baseline=(char.base)]{
            \node[shape=circle,draw,inner sep=1pt] (char) {#1};}}

\renewcommand{\footnotemark}{\textsuperscript{\arabic{footnote}}}

\def\paperID{3467} % *** Enter the Paper ID here
\def\confName{CVPR}
\def\confYear{2025}

\title{TurboFill: Adapting Few-step Text-to-image Model for Fast Image Inpainting}

\author{
  Liangbin Xie$^{1,4}$ 
  \hspace{9pt} Daniil Pakhomov$^{3}$ 
  \hspace{9pt} Zhonghao Wang$^{3}$ 
  \hspace{9pt} Zongze Wu$^{3}$ 
  \hspace{9pt} Ziyan Chen$^{4}$ \\
  Yuqian Zhou$^{3}$
  \hspace{9pt} Haitian Zheng$^{3}$ 
  \hspace{9pt} Zhifei Zhang$^{3}$ 
  \hspace{9pt} Zhe Lin$^{3}$ 
  \hspace{9pt} Jiantao Zhou$^{1}$$^\dagger$ 
  \hspace{9pt} Chao Dong$^{2,4}$$^\dagger$ \\ 
  \vspace{-0.05cm}
\small$^1$State Key Laboratory of Internet of Things for Smart City, University of Macau \hspace{3pt}
\small$^2$Shenzhen University of Advanced Technology \hspace{3pt} \\
\small$^3$Adobe\hspace{3pt}
\small$^4$Shenzhen Institutes of Advanced Technology, Chinese Academy of Sciences \\
}
\begin{document}

\twocolumn[{%
\renewcommand\twocolumn[1][]{#1}%
\maketitle
\begin{figure}[H]
\vspace{-1.3cm}
\hsize=\textwidth
\centering
% \fbox{\rule{0pt}{4in} \rule{0.9\linewidth}{0pt}}
\includegraphics[width=7 in]{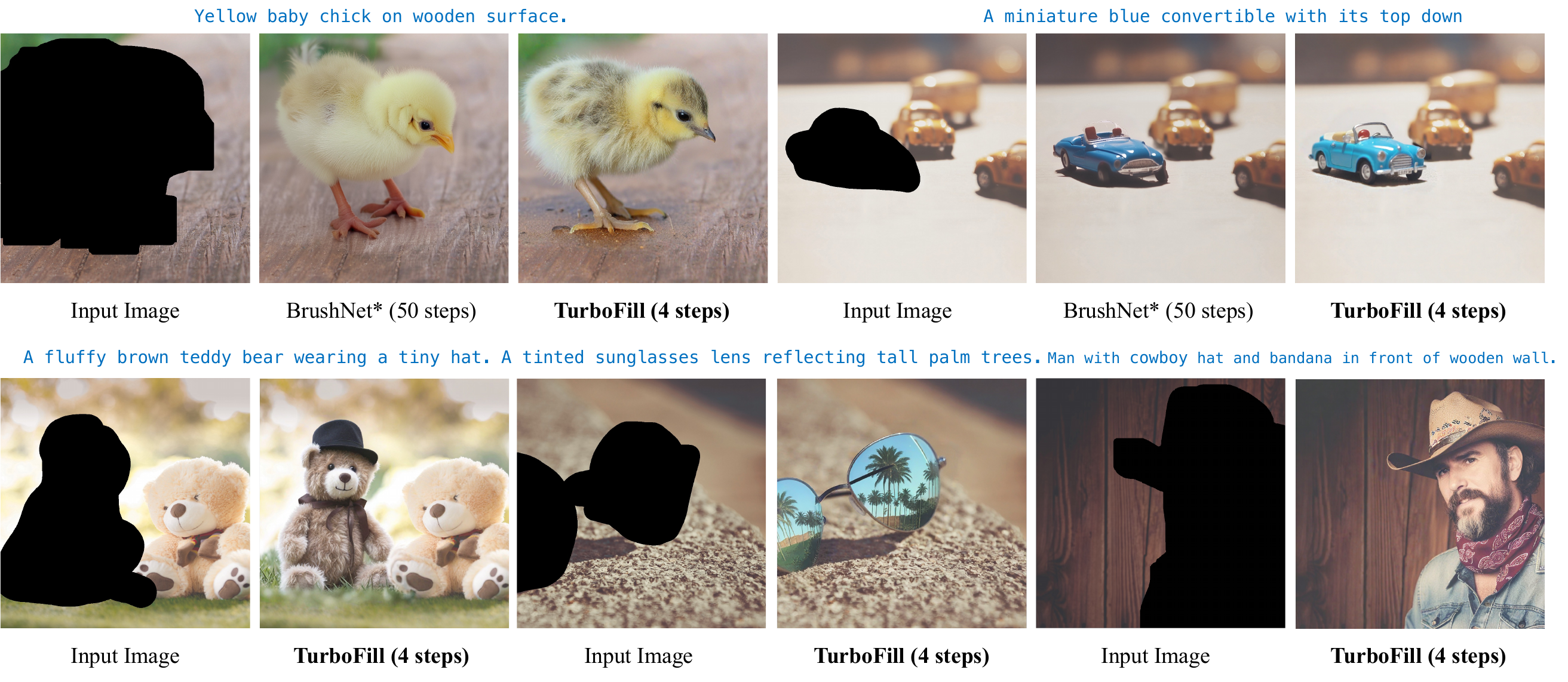}
\vspace{-0.7cm}
\caption{We propose \textbf{TurboFill}, a fast image inpainting method that leverages a 3-step adversarial training scheme. With only four diffusion steps, TurboFill outperforms the multi-step BrushNet*~\cite{ju2024brushnet}, delivering realistic details and textures with remarkable efficiency.
}
\vspace{-0.10cm}
\label{fig:real_teaser}
\end{figure}
}]
% \maketitle

% \vspace{-5cm}
\let\thefootnote\relax\footnotetext{$^\dagger$ Corresponding author}
\begin{abstract}
This paper introduces TurboFill, a fast image inpainting model that enhances a few-step text-to-image diffusion model with an inpainting adapter for high-quality and efficient inpainting. While standard diffusion models generate high-quality results, they incur high computational costs. We overcome this by training an inpainting adapter on a few-step distilled text-to-image model, DMD2, using a novel 3-step adversarial training scheme to ensure realistic, structurally consistent, and visually harmonious inpainted regions. To evaluate TurboFill, we propose two benchmarks: DilationBench, which tests performance across mask sizes, and HumanBench, based on human feedback for complex prompts. Experiments show that TurboFill outperforms both multi-step BrushNet and few-step inpainting methods, setting a new benchmark for high-performance inpainting tasks. The project page is available % at \url{https://liangbinxie.github.io/projects/TurboFill/}.
\href{https://liangbinxie.github.io/projects/TurboFill/}{here}.
\end{abstract}    
\section{Introduction}
\label{sec:intro}

\begin{figure*}[t]
\centering
\small 
\begin{minipage}[t]{1.0\linewidth}
\vspace{-1cm}
\centering
\includegraphics[width=1.0\columnwidth]{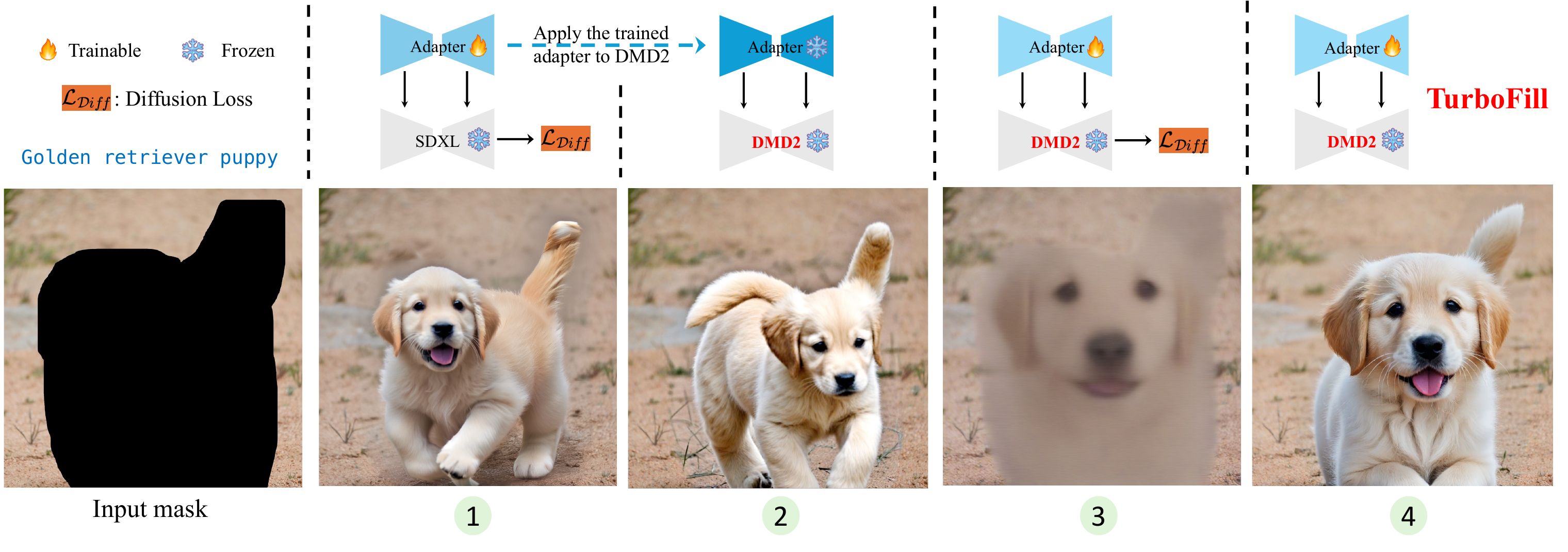}
\end{minipage}
\centering
\vspace{-0.4cm}
\caption{ (\textbf{Zoom in for best view}) 1. The multi-step adapter achieves high-quality inpainting results but incurs significant inference costs, requiring over 50 diffusion steps.
2. Applying the pre-trained multi-step BrushNet adapter directly to the few-step U-Net (DMD2) results in artifacts, including oversaturated colors and semantic inconsistencies (e.g., generating a dog with two tails).
3. Training the adapter-DMD2 solely with diffusion loss produces blurred outputs with low-quality inpainting results.
4. In contrast, training the adapter-DMD2 using the proposed 3-step adversarial training scheme yields high-quality inpainting results,  
requiring only four diffusion steps. } 
\vspace{-5pt}
\label{fig:teaser}
\end{figure*}

Image inpainting, the task of filling missing regions in an image, has seen significant advancement with the rise of deep generative models~\cite{ju2024brushnet, chen2024powerpaint, avrahami2023bld, manukyan2023hd}. Existing diffusion-based inpainting methods have shown promising results for various applications such as object removal, regeneration, and text-guided object insertion. 

The most straightforward way of training diffusion-based inpainting models is to add the masked image (background conditioning) to the input channels of a pretrained diffusion model and finetune it with inpainting data, as shown by SDXL-inpaint \cite{sdxl_inpaint}. However, such methods typically require fine-tuning the entire network, entailing large training datasets and significant computational resources. In contrast, adapter-based approaches, like BrushNet \cite{ju2024brushnet}, utilize a ControlNet \cite{zhang2023controlnet} to capture conditioning information and inject features into a frozen diffusion U-Net. BrushNet achieves state-of-the-art inpainting results while incurring relatively low training costs in terms of data and computational resources, owing to the adapter’s efficiency and the preservation of the U-Net’s original generation capacity.

Despite the advantage of BrushNet-like models, they still depend on many iterative sampling steps, leading to substantial inference cost in practice. Interestingly, we found that an inpainting adapter trained on a pretrained text-to-image diffusion model can be seamlessly integrated into its corresponding few-step student model without further training, while yielding reasonable quality results (Fig.~\ref{fig:teaser} \protect\circled{2}). This can be attributed to the semantic alignment between the teacher and student models. However, this naive migration approach exhibits quality issues such as color shifts, over-saturation, and loss of details.

Alternatively, we can use a two-stage approach by first training a teacher BrushNet and then applying advanced diffusion distillation techniques \cite{luo2023latent, sauer2023adversarial, yin2023one, lin2024sdxl, yin2024improved} to the teacher BrushNet to reduce steps while maintaining quality. Although recent distillation methods for diffusion models have demonstrated remarkable performance, we have found that due to BrushNet's large model size, it demands excessive memory and incurs high computational costs, making the training process difficult to manage.

In this paper, we propose a simpler single-stage training strategy by training an inpainting adapter directly on top of a pretrained few-step distilled text-to-image model (e.g., DMD2). We observe that naively training an inpainting adapter and few-step model solely with diffusion loss results in blurry, low-quality outputs (Fig.~\ref{fig:teaser} \protect \circled{3}). To address these shortcomings and enhance inpainting performance, we introduce a diffusion discriminator to guide optimization of the adapter-DMD2 generator. The diffusion discriminator is trained to differentiate between noisy real images and fake latents, while a combined diffusion loss ensures the model learns meaningful semantic representations.

In summary, TurboFill is trained with a novel 3-step adversarial training scheme with two generator update steps and one discriminator update step. Our model enables high-quality image inpainting with only a few diffusion steps. Compared to the naive approach of training a multi-step adapter (requiring 8 V100 GPUs for 72 hours) followed by distillation (requiring 64 A100 GPUs for 50 hours), TurboFill attains comparable outcomes using merely 8 A100 GPUs within 50 hours. This accomplishment marks a reduction in training time by over a factor of 10 and a substantial enhancement in training efficiency.

Existing benchmarks often fall short in simulating real-world user operations. To address this gap, we introduce two new benchmarks: \textbf{DilationBench} and \textbf{HumanBench}. DilationBench provides an objective measure by testing the model on the masks with varying dilation sizes, while HumanBench incorporates human user feedback and complex prompts, offering insights into real-world performance. These benchmarks contribute a robust evaluation framework that can facilitate advancements in inpainting models and more accurately reflect user-oriented performance.

Extensive quantitative and qualitative evaluations demonstrate that TurboFill not only supports complex prompts and user-driven inpainting tasks but also outperforms existing few-step inpainting models in terms of speed and quality. In summary, our contributions are as follows:

\begin{itemize}
    \item A novel 3-step adversarial training scheme is proposed to train an inpainting adapter directly on top of few-step text-to-image diffusion model, which helps the model to focus on image quality and improves textures, details, and scene harmonization. 
    \item New benchmarks, DilationBench and HumanBench, are designed to better evaluate inpainting methods in user-centered scenarios.
    \item We demonstrate that our few-step TurboFill outperforms multi-step BrushNet for the inpainting task and achieves state-of-the-art performance in a few-step setting through extensive experiments. 
\end{itemize}
\section{Related Work}

Image inpainting, a classic problem in computer vision, focuses on reconstructing missing regions in an image with realistic and natural content. Early methods, including traditional techniques, VAEs, and GANs~\cite{barnes2009patchmatch, elharrouss2020imageinpainting, cao2024leftrefill,yu2018generative,yu2019free,zeng2019learning,zeng2022aggregated}, often struggled to produce high-quality results. Recently, diffusion models have gained popularity for their ability to generate high-quality, diverse outputs with fine-grained control, demonstrating significant advancements over earlier approaches ~\cite{ju2024brushnet, chen2024powerpaint, avrahami2023bld, manukyan2023hd}. 
Diffusion models, particularly the Stable Diffusion family, have introduced a robust framework for image generation and inpainting by iteratively denoising latent representations. Stable Diffusion-based methods have achieved impressive results in high-resolution image inpainting, yet are computationally intensive due to multi-step sampling. 
% Based on diffusion models, 

Inpainting tasks based on diffusion models generally follow two approaches: training an entire diffusion model or incorporating external modules. The first approach, like PowerPaint~\cite{chen2024powerpaint}, CATDiffusion~\cite{chen2025catdiffusion}, and SDXL-Inpainting, focuses on adapting the model itself. PowerPaint introduces learnable task prompts along with tailored fine-tuning strategies to guide the model's focus on specific inpainting targets, while SDXL-Inpainting~\cite{sdxl_inpaint} directly utilizes both the background region and the mask as inputs to fine-tune a new model. In contrast, the second approach, exemplified by methods such as ControlNet~\cite{zhang2023controlnet} or T2I-Adapter~\cite{mou2024t2i}, leverages external modules to enhance the diffusion model's capabilities without retraining it entirely. Recent methods, such as BrushNet~\cite{ju2024brushnet}, employ mask-specific prompts, refining the generated content within the masked area for improved coherence with the background.

Our method builds on these developments by leveraging a ControlNet-like adapter for DMD2, integrating both GAN and diffusion losses to enhance image realism and style coherence. By focusing on masked-region-specific prompts and incorporating fine-grained diffusion control, our approach achieves fast and high-quality inpainting results that align closely with user inputs and surrounding context, advancing upon prior work in speed, quality, and flexibility in real-world applications.

\section{Preliminary}

% \subsection{Stable Diffusion}

The forward diffusion process gradually adds Gaussian noise $\epsilon_t$ to a clean image $x_0$ by 

\begin{equation} \label{equation:alpha}
x_t=\sqrt{\bar{\alpha}_t}x_0+\sqrt{1-\bar{\alpha}_t}\epsilon_t,
\end{equation}

\noindent where $\bar{\alpha}_t$ controls noise schedule. In the training phase, a denoising network $\epsilon_\theta$ is trained to predict $\epsilon_t$ given $x_t$, timestep $t$ and text prompt $c$, with the diffusion loss:
\begin{equation} \label{equation:sampling}
L(\hat{\epsilon}_\theta)=\mathbb{E}_{x_0 \sim q;\epsilon_t \sim \mathcal{N}(0,1)}[\parallel \epsilon_\theta(x_t, c, t) - \epsilon_t \parallel^2].
\end{equation}

To enable condition generation, adapter-based methods \cite{zhang2023adding, mou2024t2i, ju2024brushnet} introduce additional conditioning information via a lightweight adapter, and achieve efficient training (since the main diffusion network remains frozen), with a slight inference overhead. 

Traditionally, reverse diffusion processes necessitate a computationally intensive multi-step iterative sampling procedure (typically 20-50 steps). Distillation techniques \cite{luo2023latent, sauer2023adversarial, yin2023one, lin2024sdxl} have been developed to reduce this requirement, achieving accelerated sampling with as few as 1-4 steps. However, these distillation processes are resource-intensive in both computational power and time. In contrast, our approach directly trains the adapter with both the multi-step and few-step text-to-image diffusion model, eliminating the need for a separate distillation process, thereby enabling more efficient training and inference.

\section{Our Approach}

\begin{figure*}[t]
\centering
\small 
\begin{minipage}[t]{\linewidth}
\vspace{-1cm}
\centering
\includegraphics[width=0.95\columnwidth]{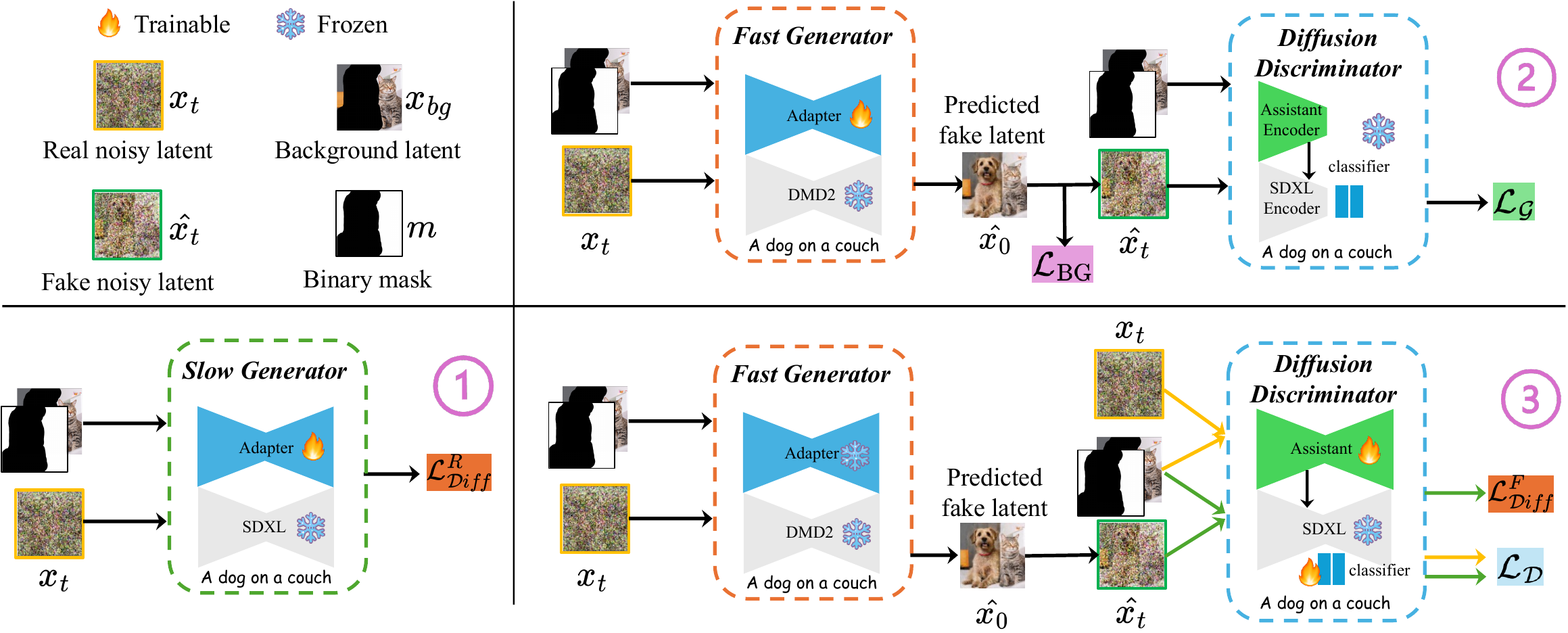}
\end{minipage}
\centering
\vspace{-0.2cm}
\caption{The training of TurboFill alternates between 3 steps: \protect \circled{1}. optimizing the adapter using the gradient of $\mathcal{L}^{R}_{\mathrm{Diff}}$, and  \protect \circled{2}. $\mathcal{L}_{\mathcal{G}}$ and $\mathcal{L}_{\mathrm{BG}}$ are employed to update the adapter in fast generator, and \protect \circled{3}. $\mathcal{L}^{F}_{\mathrm{Diff}}$ and $\mathcal{L}_{\mathcal{D}}$ are jointly applied to update the parameters of the diffusion discriminator module. 
%These three steps are repeatedly cycled throughout the training process. 
Note that the adapters in slow generator and fast generator share same weights during training.}
\vspace{-10pt}
\label{fig:framework} 
\end{figure*}

In this work, we propose a fast inpainting model, \textbf{TurboFill}, which directly trains an inpainting adapter on top of a few-step text-to-image diffusion model. 
As shown in Fig.~\ref{fig:framework}, there are three components: the fast generator, the slow generator and the diffusion discriminator. All these three components aim to enhance the adapter's capability of inpainting masked images. Specifically, the slow generator helps the adapter to denoise the noisy latents towards the most probable direction of realistic image generation. The fast generator enables the adapter to generate a clean image for critique during training. The diffusion discriminator guides the adapter to generate images with better textures, details and content harmonization. Based on these three components, TurboFill follows a 3-step adversarial training scheme. The training alternates between 3 steps throughout the training process. In the following sections, we provide a detailed description of the training procedure for each step and the components.

\subsection{Training the Adapter in Slow Generator}
To capture conditioning information and introduce additional features to the diffusion model, we utilize an inpainting adapter with the same ControlNet architecture as BrushNet~\cite{ju2024brushnet}. Given that SDXL is a multi-step text-to-image generator, we designate the integration of the inpainting adapter and SDXL as the `slow generator'. In contrast, DMD2 is a few-step text-to-image generator, so the combination of the inpainting adapter and DMD2 is named the `fast generator'. 
In both generators, conditioning information is introduced by concatenating the noisy latent $x_t$, the background image latent $x_{bg}$, and a downsampled binary mask $m$. This concatenated input is then processed by the inpainting adapter, while $x_t$ is simultaneously fed into the UNet. Then, the features produced by the inpainting adapter are fused with the UNet's features via residual connection.

As shown in Fig.~\ref{fig:framework} \circled{1}, the parameters of the inpainting adapter in slow generator are updated based on the real diffusion loss:
\begin{align}
\mathcal{L}^{R}_{\text{Diff}}=\min_{\epsilon_\theta}\mathbb{E}_{x_t, t \sim [0, T], \epsilon \sim \mathcal{N}(0,1)} \\
\| \epsilon \nonumber - \epsilon_\theta^{\scriptstyle \text{SDXL}} \left( x_t, [x_{bg}, x_t, m], t \right) \Big\|_2^2,
% \tag{1}
\label{eq:diffusion}
\end{align}
where $T=1,000$ and DDPM scheduler~\cite{ho2020denoising} is employed to generate $x_t$.

\subsection{Training the Adapter in Fast Generator}
Given that adversarial training is highly effective in minimizing the divergence between different distributions, we employ adversarial training to minimize the discrepancy between real latent distribution and the fake latent distribution $\{\hat{x}_{0}\}$. Specifically, we first generate the clean fake latent $\hat{x_0}$ from the noisy latent $x_t$ with the fast generator in a single step, as described by 
\begin{equation}
\hat{x}_{0}=\frac{x_t-\sqrt{1-\bar{\alpha}_t} \times \epsilon_\theta^{\scriptstyle \text{DMD2}}\left(x_t, [x_{bg}, x_{t}, m], t\right)}{\sqrt{\bar{\alpha}_t}},
\label{equ:dmd2}
\end{equation}
where $\bar{\alpha}_t$ is determined by the LCM scheduler~\cite{luo2023latent} and $t$ is the sampled timestep. 
Then, the noisy version $\hat{x_t}$ originated from $\hat{x_0}$ is fed into the diffusion discriminator. Meanwhile, to preserve the background region, $\hat{x_0}$ is used to compute the background loss. 
 
\noindent \textbf{Diffusion discriminator.} As shown in Fig.~\ref{fig:framework} \circled{2}, the diffusion discriminator $\mathcal{D}_\phi$ consists of the SDXL encoder, the assistant encoder, and the convolution-based classifier. For assistant encoder, we use the same encoder structure of inpainting adapter. It takes concatenated noisy latent, downsampled binary mask, and masked image latent $x_{bg}$ as the input. The features extracted by the assistant encoder are progressively integrated into the SDXL encoder. The final feature maps are then passed through the classifier and mapped to a one-dimensional vector. Based on the vector, we calculate $\mathcal{L}_{\mathcal{G}}$:
\begin{equation}
\mathcal{L}_{\mathcal{G}}=\min_{\epsilon_\theta} -\mathbb{E}_{y_{t}, t \sim[0, T]}\left[\log \mathcal{D}_\phi\left(\hat{x}_{0}, [x_{bg}, \hat{x}_{0}, m], t)\right)\right],
\end{equation}
where $T=1,000$.

\noindent \textbf{Background preservation.} For the image inpainting task, it is essential to preserve the background region. Therefore, for the clean fake latent $\hat{x_{0}}$, we apply a reconstruction loss to the background region to minimize the distance between $\hat{x_{0}}$ and the clean latent $x_{0}$:

\begin{equation}
\mathcal{L}_{\mathrm{BG}}=
\left\| x_0 \odot m - \hat{x}_{0} \odot m \right\|_2^2,
\end{equation}
where $\odot$ represents point-wise multiplication. Based on $\lambda_{1}\mathcal{L}_{\mathcal{G}}+\lambda_{2}\mathcal{L}_{\mathrm{BG}}$, in this step, only the parameters of the inpainting adapter are updated.

\subsection{Training the Diffusion Discriminator}

When updating the discriminator, we fix the parameters of the inpainting adapter and update the parameters of the assistant encoder and classifier. Specifically, both the real noisy latent $x_{t}$ and the fake noisy latent $\hat{x_{t}}$ are input into the diffusion discriminator to compute the loss $\mathcal{L}_{\mathcal{D}}$:
\begin{equation}
\begin{aligned}
\mathcal{L}_{\mathcal{D}} = &- \mathbb{E}_{\hat{x}_{0, t}, t \sim[0, T]} \left[ \log (1 - \mathcal{D}_\phi\left(\hat{x}_{0}, [x_{bg}, \hat{x}_{0}, m], t)\right)) \right] \\
&- \mathbb{E}_{x_{t}, t \sim[0, T]} \left[ \log \mathcal{D}_\phi\left(x_t, [x_{\mathrm{bg}}, x_{t}, m], t)\right) \right].
\end{aligned}
\end{equation}

Although the inpainting adapter optimized only with GAN loss can generate images, the generated images are of inferior quality and commonly incorporate unwanted elements. We attribute this issue to the limitations of the GAN loss, which rarely assists the discriminator in comprehending the image structure. In contrast, diffusion loss excels at scene understanding. Therefore, we introduce the fake diffusion loss to train the discriminator while allowing the GAN loss to focus more on enhancing image textures and object details. To compute the diffusion loss for the fake noisy latent $\hat{x_t}$, we introduce an assistant decoder, following the assistant encoder. The fake diffusion loss is:
\begin{align}
\mathcal{L}^{F}_{\mathrm{Diff}}=\min_{\mathcal{A}_\psi}\mathbb{E}_{y_t, t \sim [0, T], \epsilon \sim \mathcal{N}(0,1)} \\
\| \epsilon \nonumber - \mathcal{A}_\psi^{\scriptstyle \text{SDXL}} \left( x_t, [x_{bg}, \hat{x}_{t}, m], t \right) \Big\|_2^2,
% \tag{1}
\label{eq:diffusion}
\end{align}
where $\mathcal{A}_\psi$ represents the parameters of assistant encoder and decoder. The combination of $\mathcal{L}_{\text{Diff}}^F+\lambda_3 \mathcal{L}_{\mathcal{D}}$ is used as the loss to optimize the assistant encoder, assistant decoder modules, and classifier.
\section{Experiments}

\subsection{Datasets and Metrics}

\subsubsection{Training Dataset}
To enable editing through the input of a local mask and the corresponding prompt for the masked region, we construct a dataset that includes local masks and their associated textual descriptions. Specifically, we crawl approximately $1,200,000$ images from Internet and use Florence-2~\cite{xiao2024florence} to perform the dense region caption task with the prompt \textless DENSE\_REGION\_CAPTION\textgreater. This task identifies the main objects in the images and generates their corresponding short descriptions. These descriptions are then input into SAM2~\cite{ravi2024sam} to obtain the segmentation masks for the objects. We name this dataset as LocalCaptionData.

\subsubsection{Evaluation Datasets}

The commonly used public datasets in the field of image inpainting are primarily OpenImage V6~\cite{kuznetsova2020openimage}, MSCOCO~\cite{lin2014microsoft}, and BrushBench~\cite{ju2024brushnet}. Among these, OpenImage V6 and MSCOCO provide precise segmentation masks and bounding box masks, which are used to perform inpainting on the masked regions based on class names. BrushBench only offers accurate segmentation masks, where inpainting is performed on the masked regions using a global image description. However, considering that users typically perform inpainting with rough masks and corresponding descriptions (which may sometimes be complex and contain multiple attributes), these three datasets significantly diverge from real-world scenarios. To better simulate user operations, we construct two evaluation datasets:

\begin{itemize}
    \item \textbf{DilationBench}: The masks are created by randomly dilating the segmentation masks, and the prompts describe the content within the masked area of the original image. DilationBench contains 300 pairs of masks and prompts.
    \item \textbf{HumanBench}: The masks are manually labeled, and the prompts are manually written. HumanBench contains 150 pairs of masks and prompts.
\end{itemize} 
The descriptions and visualizations of the two benchmark datasets are included in the supplementary material. 

\subsubsection{Evaluation Metrics}
To evaluate the quality of inpainted images, we use three metrics: Q-Align~\cite{wu2023qalign}, CLIPIQA+~\cite{wang2023clipiqa}, TOPIQ~\cite{chen2024topiq}. They assess visual quality from different perspectives. Q-Align, which leverages a large language model (LLM) to generate visual scores, is state-of-the-art on various Image Quality Assessment (IQA) leaderboards\textsuperscript{1}\footnote{\href{https://iqa-pytorch.readthedocs.io/en/latest/benchmark.html}{https://iqa-pytorch.readthedocs.io/en/latest/benchmark.html}}. Excluding Q-Align, CLIPIQA+ and TOPIQ rank first in non-reference and image aesthetic benchmarks, respectively, making them reliable indicators of visual quality and providing complementary insights. For a more comprehensive evaluation of inpainting results, we consider not only the masked region but also the entire image, ensuring that both local quality in the inpainted area and overall visual coherence across the entire image are assessed. To measure the consistency between the text and the content generated in the mask region, we use the CLIP Similarity metric~\cite{radford2021clip}. Higher values for all four metrics indicate better performance.

\begin{figure*}[t]
\centering
\small 
\begin{minipage}[t]{0.95\linewidth}
\vspace{-1cm}
\centering
\includegraphics[width=1.0\columnwidth]{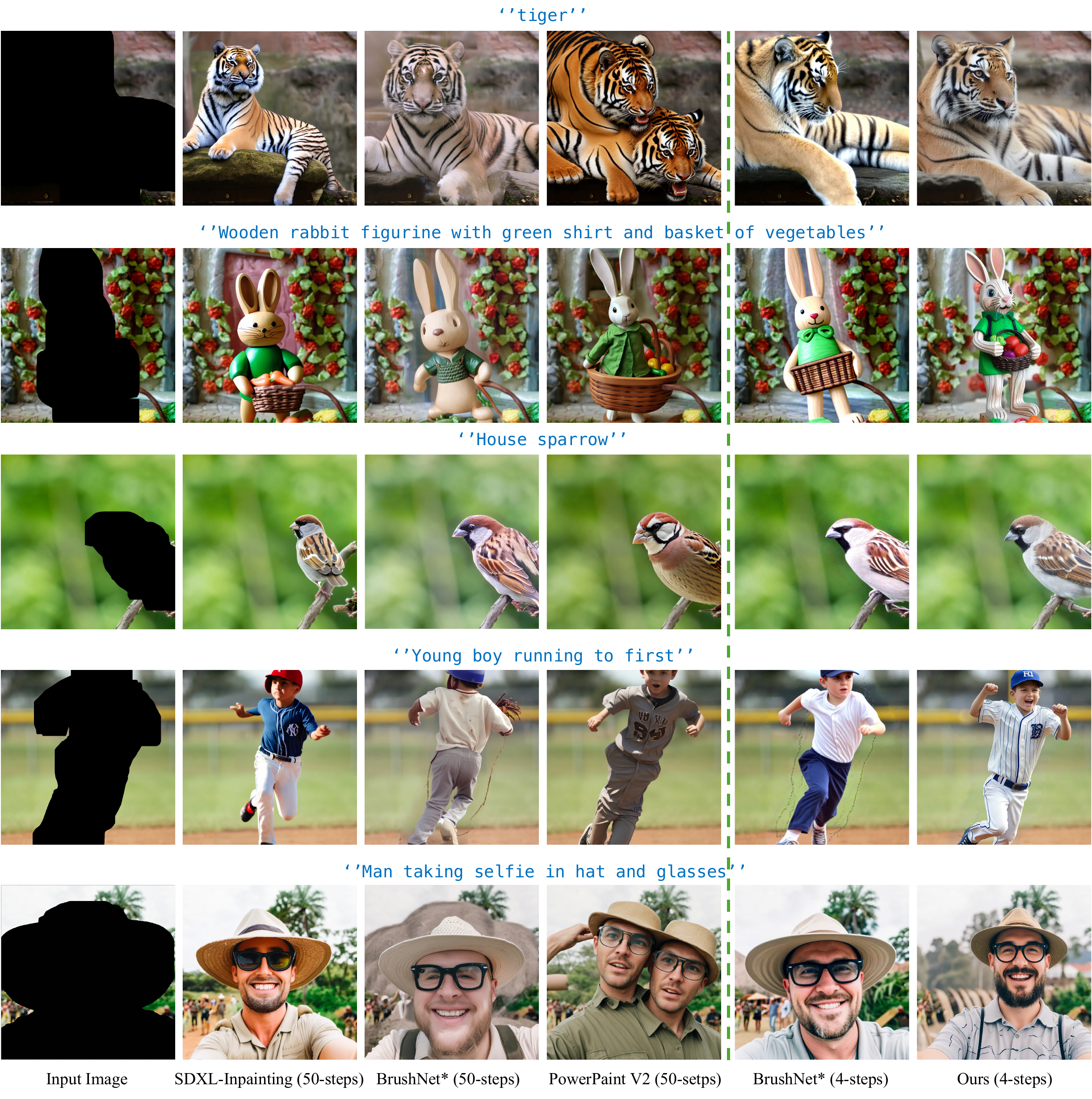}
\end{minipage}
\centering
\vspace{-0.2cm}
\caption{Comparison of previous inpainting methods and BrushNet on DilationBench. Compared to other methods, TurboFill generates more realistic details and textures in just 4 steps, while achieving good scene harmonization. (\textbf{Zoom in for best view})}
\vspace{-10pt}
\label{fig:qualitative_comparison}
\end{figure*}

\subsection{Experimental Settings}
\subsubsection{Network Configurations} Inpainting adapter adopts the same architecture as BrushNet, with modifications to the SDXL model. Specifically, the self-attention and cross-attention modules are removed. Note that our approach does not impose any architectural constraints on inpainting adapter. Assistant also adopts U-Net like structure, for simiplity, we directly adopt the same architecture as inpainting adapter. The classifier is composed of convolutional layers, GroupNorm~\cite{wu2018group}, and SiLU~\cite{elfwing2018silu} activation, which together map the feature map to a 1-dimensional vector. The details of classifier are described in the supplementary material.

When using SDXL as the base model for noise prediction, we employ a DDIM scheduler~\cite{song2021denoising} and sample randomly from 1,000 timesteps. In contrast, when using DMD2 as the base model, we use an LCM scheduler~\cite{luo2023latent} and randomly sample from four specific timesteps: $\{999, 749, 499, 249\}$.

\subsubsection{Implementation Details}

All experiments are performed on 8 A100 GPUs with 80GB memory. During training, the learning rate for parameter updates is set to 1e-5, with a batch size of 2 and gradient accumulation steps configured to 4. We adopt mixed precision to optimize computational efficiency. $\lambda_{1}, \lambda_{2}, \lambda_{3}$ are set to $1e^{-3}$, $1e^{-1}$, and $1e^{-2}$, respectively. All models are trained for 40K iterations using the AdamW optimizer.

\subsection{Quantitative Comparisons}
We compare TurboFill with existing state-of-the-art image inpainting methods, namely BLD~\cite{avrahami2023bld}, HD-Painter~\cite{manukyan2023hd}, SDXL-Inpainting\textsuperscript{1}\footnote{\href{https://huggingface.co/diffusers/stable-diffusion-xl-1.0-inpainting-0.1}{https://huggingface.co/diffusers/stable-diffusion-xl-1.0-inpainting-0.1}}, BrushNet-Rand~\cite{ju2024brushnet}, and PowerPaint~\cite{chen2024powerpaint}. All of these methods follow a multi-step approach, requiring $50$ steps during inference to generate inpainting results. While BrushNet-Rand, trained with random masks, effectively handles imprecise masks and uses a full image description during inference, the other methods rely only on descriptions of the masked region.

% \usepackage{multirow}
\begin{table*}[]
\vspace{-1cm}
\centering
\resizebox{0.95\textwidth}{!}{
\begin{tabular}{cc|ccc|ccc|c}
\hline
\multicolumn{2}{c|}{Metrics}                                                 & \multicolumn{3}{c|}{Mask Region Quality}            & \multicolumn{3}{c|}{Whole Image Quality}            & Text-Align      \\ \hline
\multicolumn{2}{c|}{Method}                                                  & Q-Align         & CLIPIQA+        & TOPIQ           & Q-Align         & CLIPIQA         & TOPIQ           & CLIP Sim        \\ \hline
\multicolumn{1}{c|}{\multirow{6}{*}{MultiStep (50 steps)}} & BLD             & 4.1836          & 0.6874          & 5.1781          & 4.3320          & 0.6791          & 5.4662          & 24.884          \\
\multicolumn{1}{c|}{}                                      & HD-Painter      & 3.9727          & 0.6611          & 4.9345          & 4.4805          & 0.6439          & 5.4566          & 25.585 \\
\multicolumn{1}{c|}{}                                      & SDXL-Inpainting &    4.2461    & 0.6666          & 5.1533          &   4.6172      & 0.6695          & 5.5409          & 24.848          \\
\multicolumn{1}{c|}{}                                      & BrushNet-Rand   & 4.2617          & 0.6729          & 5.1738          & 4.6289          & 0.6818          & 5.5779 & 21.634          \\
\multicolumn{1}{c|}{}                                      & BrushNet*       & 4.4688          & 0.7137          & 5.2504          & 4.5312          & 0.6974          & 5.5585          & 25.389 \\
\multicolumn{1}{c|}{}                                      & PowerPaint V2     & \textbf{\textcolor{red}{4.7773}} & \textbf{\textcolor{red}{0.7765}} & \textbf{\textcolor{red}{5.5468}} & \textbf{\textcolor{red}{4.7227}} & \textbf{\textcolor{red}{0.7371}} & \textbf{\textcolor{red}{5.6474}} & \textbf{\textcolor{red}{26.256}} \\ \hline
\multicolumn{1}{c|}{\multirow{3}{*}{FewStep (4 steps)}}    & BrushNet-Rand   & 4.1602          & 0.6538          & 4.9927          & 4.5547          & 0.6608          & 5.4599          & 21.831          \\
\multicolumn{1}{c|}{}                                      & BrushNet*  & 4.1836          & 0.6572          & 4.9939          & 4.4492          & 0.6427          & 5.3961          & 25.341          \\
\multicolumn{1}{c|}{}                                      & TurboFill            & \textbf{\textcolor{blue}{4.5703}} & \textbf{\textcolor{blue}{0.7332}} & \textbf{\textcolor{blue}{5.2753}} & \textbf{\textcolor{blue}{4.7188}} & \textbf{\textcolor{blue}{0.7111}} & \textbf{\textcolor{blue}{5.5392}} & \textbf{\textcolor{blue}{25.352}}          \\ \hline
\end{tabular}
}
\vspace{-5pt}
\caption{Quantitative comparisons among TurboFill and other diffusion based inpainting models in DilationBench. \textbf{\textcolor{red}{Red}} and \textbf{\textcolor{blue}{blue}} indicates the best multi-step and the best few-step performances, respectively.}
\label{tab:dilationbench}
\end{table*}

% \usepackage{multirow}
\begin{table*}[]
\centering
\resizebox{0.95\textwidth}{!}{
\begin{tabular}{cc|ccc|ccc|c}
\hline
\multicolumn{2}{c|}{Metrics}                                                 & \multicolumn{3}{c|}{Mask Region Quality}            & \multicolumn{3}{c|}{Whole Image Quality}            & Text-Align      \\ \hline
\multicolumn{2}{c|}{Method}                                                  & Q-Align         & CLIPIQA+        & TOPIQ           & Q-Align         & CLIPIQA         & TOPIQ           & CLIP Sim        \\ \hline
\multicolumn{1}{c|}{\multirow{5}{*}{MultiStep (50 steps)}} & BLD             & 4.1523          & 0.6908          & 5.2996          & 3.9062          & 0.6579          & 5.4199          & 24.767          \\
\multicolumn{1}{c|}{}                                      & HD-Painter      & 3.9844          & 0.6493          & 5.0873          & 4.2070          & 0.6119          & 5.4381          & 25.996 \\
\multicolumn{1}{c|}{}                                      & SDXL-Inpainting & 3.9551          & 0.6405          & 5.2040          & 4.0469          & 0.6288          & 5.4565          & 24.041          \\
\multicolumn{1}{c|}{}                                      & BrushNet*       & 4.2578          & 0.7063          & 5.4340 & 4.0898          & 0.6694          & 5.5480 & 25.366          \\
\multicolumn{1}{c|}{}                                      & PowerPaint V2     & \textbf{\textcolor{red}{4.5586}} & \textbf{\textcolor{red}{0.7529}} & \textbf{\textcolor{red}{5.5475}} & \textbf{\textcolor{red}{4.3633}} & \textbf{\textcolor{red}{0.7112}} & \textbf{\textcolor{red}{5.6338}} & \textbf{\textcolor{red}{26.264}} \\ \hline
\multicolumn{1}{c|}{\multirow{2}{*}{FewStep (4 steps)}}    & BrushNet*       & 4.0508          & 0.6327          & 5.1041          & 4.1484          & 0.6003          & 5.3505          & 25.473          \\
\multicolumn{1}{c|}{}                                      & Ours            & \textbf{\textcolor{blue}{4.4727}} & \textbf{\textcolor{blue}{0.7257}} & \textbf{\textcolor{blue}{5.3865}}          & \textbf{\textcolor{blue}{4.3203}} & \textbf{\textcolor{blue}{0.6822}} & \textbf{\textcolor{blue}{5.4992}}          & \textbf{\textcolor{blue}{25.710}}          \\ \hline
\end{tabular}
}
\vspace{-5pt}
\caption{Quantitative comparisons among TurboFill and other diffusion based inpainting models in HumanBench. \textbf{\textcolor{red}{Red}} and \textbf{\textcolor{blue}{blue}} indicates the best multi-step and the best few-step performances, respectively.}
\vspace{-15pt}
\label{tab:humanbench}
\end{table*}

Since we adopt the architecture design of BrushNet, for fair comparison, we also train BrushNet using LocalCaptionData, referring to this configuration as BrushNet*. Notably, BrushNet, as an external module,  %integrated within the text-to-image diffusion model architecture, 
can be directly adapted into a 4-step diffusion model (e.g., DMD2) to form a 4-step BrushNet configuration. In contrast, methods such as HD-Painter and SDXL-Inpainting, which modify the base model structure, cannot integrate with a 4-step diffusion model. PowerPaint V2, an advanced version of BrushNet and SD1.5, incorporates learnable task prompts and specialized fine-tuning strategies across four distinct inpainting tasks. However, since PowerPaint lacks an SDXL version, it cannot be adapted to a 4-step DMD2 model.

Comparative results for these methods on both DilationBench and HumanBench datasets are presented in Tab.~\ref{tab:dilationbench} and Tab.~\ref{tab:humanbench}. Given that HumanBench, like real inpainting tasks, does not provide a global description of the entire image, we exclude BrushNet-Rand from evaluation on this dataset. From Tab.~\ref{tab:dilationbench}, we find that BrushNet*, trained with LocalCaptionData, shows improvements of 3.755 and 3.52 in the text-align metric for the 50-step and 4-step models, respectively, compared to BrushNet. This suggests that training with segmentation masks and paired local descriptions, rather than global descriptions, significantly enhances BrushNet's text alignment capability.

Additionally, we observe that directly inserting BrushNet-Rand and BrushNet* into DMD2 to form a 4-step model results in no significant change in text alignment performance. However, the image quality of the final inpainting results decreases noticeably. This suggests that while replacing SDXL with a few-step model (DMD2) can accelerate BrushNet, it leads to a reduction in quality. In contrast, TurboFill can effectively mitigates this issue, showing significant improvements across all three metrics—mask region quality, whole image quality, and text alignment. Even with 4 steps, our method outperforms the 50-step BrushNet-Rand and BrushNet* in many image quality metrics (e.g., Q-Align, CLIPIQA+). This phenomenon can be observed on both benchmarks.

PowerPaint V2 currently achieves the best results across these metrics. When comparing our method with PowerPaint, we find that the gap in whole image quality is smaller than in mask image quality. Visual comparisons (Fig.~\ref{fig:qualitative_comparison}) reveal that although PowerPaint V2 generates rich textures, the results in the mask region are less natural, with noticeable style mismatch compared to the background region. This inconsistency leads to a lower performance in whole image quality compared to mask region quality. In contrast, our method produces more coherent results.
%matching the background style. 
Since no existing metrics effectively capture these differences, we conduct a user study, detailed in the supplementary file.

% \usepackage{multirow}
% \usepackage{multirow}
\begin{table*}[]
\centering
\vspace{-1cm}
\resizebox{0.72\textwidth}{!}{
\begin{tabular}{c|ccc|ccc|c}
\hline
\multirow{2}{*}{Settings} & \multicolumn{3}{c|}{Mask Region Quality}            & \multicolumn{3}{c|}{Whole Image Quality}            & Text-Align \\ \cline{2-8} 
                          & Q-Align         & CLIPIQA+        & TOPIQ           & Q-Align         & CLIPIQA+        & TOPIQ           & CLIP Sim   \\ \hline
\textbf{TurboFill}                 & \textbf{4.5703} & \textbf{0.7332} & \textbf{5.2753} & \textbf{4.7227} & \textbf{0.7111} & \textbf{5.5392} & \textbf{25.352}     \\
$- \mathcal{L}_{\mathrm{BG}}$                       & 4.3672 & 0.6858 & 5.0262 & 4.5469 & 0.6570 & 5.3826 & 25.335     \\
$- \mathcal{L}^{F}_{\mathrm{Diff}}$                     & 4.1875          & 0.6554          & 4.8695          & 4.4922          & 0.6426          & 5.3297          & 25.062     \\
$- \mathcal{R}^{F}_{\mathrm{Diff}}$             & 4.0664          & 0.6322          & 4.8497          & 4.4492          & 0.6412          & 5.3062          & 24.875     \\ \hline
\end{tabular}
}
\vspace{-5pt}
\caption{The effectiveness of different losses. From top to down, we progressively remove specific losses.} 
\vspace{-10pt}
\label{tab:ablation}
\end{table*}

\vspace{-1mm}
\subsection{Qualitative Comparison}

The qualitative comparison on DilationBench is shown in Fig.~\ref{fig:qualitative_comparison}. The results from SDXL-Inpainting exhibit a yellowish color bias (e.g., tiger, man) and lack detail (e.g., boy, figurine). For PowerPaint V2, the inpainted results are very sharp in texture and rich in color, producing high scores on most IQA metrics. However, these results suffer from severe distortions (e.g., a tiger with two heads, a person with two heads, overly elongated legs) and appear less realistic and natural (e.g., third row). The results generated by BrushNet* exhibit significant artifacts, such as the tiger's fur (1st) and the man's beard (5th), as well as the appearance of strange backgrounds (5th). When BrushNet* is inserted into the 4-step DMD2 model, we observe that the results lack significant detail (e.g., 5th row) and exhibit issues with oversaturation (e.g., tiger, figurine). Compared to these methods, TurboFill generates results with higher quality in visual comparisons. While preserving detail, our results do not exhibit the oversharpness seen in PowerPaint V2 (e.g., tiger), nor the oversaturation issue observed in BrushNet* (4-steps) (e.g., rabbit). The generated content is more realistic (e.g., bird, figurine), leading to a more harmonious image overall.

\subsection{Effectiveness of LocalCaptionData}

\begin{figure}[t]
\centering
\small 
\begin{minipage}[t]{1.0\linewidth}
\centering
\includegraphics[width=0.98\columnwidth]{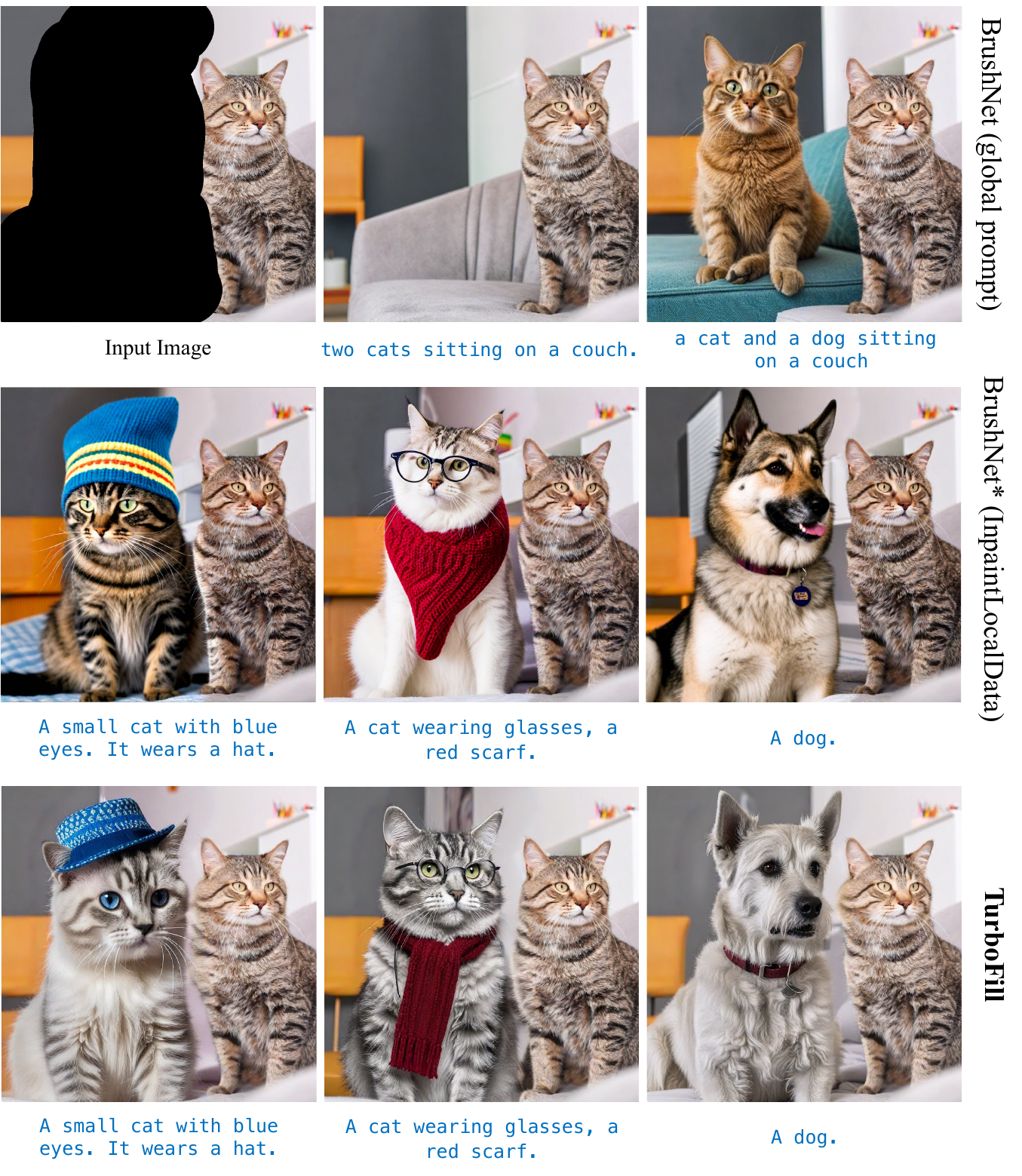}
\end{minipage}
\centering
\vspace{-10pt}
\caption{The effectiveness of LocalCaptionData. All results are obtained based on 4-step DMD2. (\textbf{Zoom in for best view})} 
%Here we set seed to $42$ for all cases.}
\vspace{-10pt}
\label{fig:inpaintlocal}
\end{figure}

BrushNet, when trained with a full image description (i.e., global prompt) as input to the base model, exhibits relatively weak text alignment capabilities, as shown in Fig.~\ref{fig:inpaintlocal}. In cases where BrushNet attempts to generate content similar or semantically related to existing objects in the image, it often either defaults to complete the background or shifts focus to other categories in the prompt (e.g., a cat). Furthermore, relying on a global prompt limits BrushNet’s effectiveness in handling more complex prompts. By using LocalCaptionData, BrushNet* effectively overcomes these limitations without any changes to the training approach, resulting in a significant improvement in text alignment performance. Additionally, TurboFill demonstrates superior performance in generating fine details.

% \vspace{+2mm}

% \vspace{-2mm}

\begin{figure}[t]
\centering
\small 
\begin{minipage}[t]{1.0\linewidth}
\centering
\includegraphics[width=1.0\columnwidth]{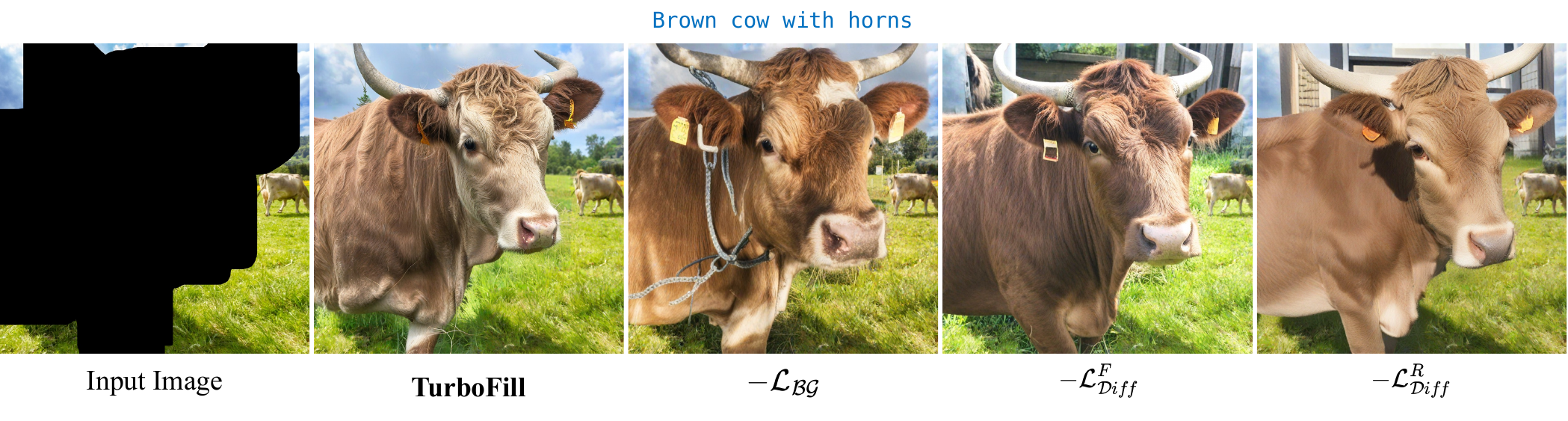}
\end{minipage}
\centering
\vspace{-10pt}
\caption{The effectiveness of different losses. From left to right, we progressively remove specific losses. (\textbf{Zoom in for best view})} 
%Here we set seed to $42$ for all cases.}
\vspace{-5pt}
\label{fig:ablation}
\end{figure}

\subsection{Ablation Studies.}

Starting from TurboFill, we remove $\mathcal{L}_{\mathrm{BG}}$, $\mathcal{L}^{F}_{\mathrm{Diff}}$, and $\mathcal{L}^{R}_{\mathrm{Diff}}$ in sequence, with quantitative and qualitative results shown in Fig.~\ref{fig:ablation} and Tab.~\ref{tab:ablation}. When $\mathcal{L}_{\mathrm{BG}}$ is removed, there is a clear performance drop in all metrics. In the visualizations, we see that without $\mathcal{L}_{\mathrm{BG}}$, the color of background region changes noticeably, creating a sharp boundary between the fill-in and background regions. Further removing $\mathcal{L}^{F}_{\mathrm{Diff}}$ introduces conflicting elements (i.e., house) in the fill-in region, suggesting the discriminator fails to fully capture the holistic scene. Finally, without $\mathcal{L}^{R}_{\mathrm{Diff}}$, relying only on GAN loss, the inpainted images exhibit not only inconsistencies with the background but also poor texture and detail. This highlights that GAN loss alone struggles to close the gap between fake and real latents. Only when combining $\mathcal{L}^{F}_{\mathrm{Diff}}$, $\mathcal{L}^{R}_{\mathrm{Diff}}$, and $\mathcal{L}_{\mathrm{BG}}$ in GAN training does the model achieve enhanced texture, detail, and effective scene harmonization between fill-in and background regions.

\vspace{-2mm}
\section{Conclusion}

In this work, we introduce \textbf{TurboFill}, a fast inpainting model designed to address the computational and quality challenges in diffusion-based inpainting tasks. By directly training an inpainting adapter on a few-step distilled text-to-image diffusion model, TurboFill achieves state-of-the-art inpainting results with significantly smaller inference costs. The proposed three-step adversarial training scheme, incorporating both GAN and diffusion losses, ensures the generation of realistic and harmonious content with fine textures and structural coherence. To evaluate TurboFill's performance, we introduce two new benchmarks, \textbf{DilationBench} and \textbf{HumanBench}, which provide robust evaluations across varying mask complexities and user-centric scenarios. Experimental results demonstrate TurboFill’s ability to outperform existing few-step and multi-step BrushNet,
%in terms of speed, visual quality, and text alignment, 
setting a new benchmark for practical and efficient inpainting.

\vspace{-2mm}
\section{Acknowledgement}
\label{sec:Acknowledgement}
% \vspace{-2mm}
This work was supported in part by Macau Science and Technology Development Fund under SKLIOTSC-2021-2023, 0022/2022/A1, and 0014/2022/AFJ; in part by Research Committee at University of Macau under MYRG-GRG2023-00058-FST-UMDF; in part by the Guangdong Basic and Applied Basic Research Foundation under Grant 2024A1515012536; in part by National Natural Science Foundation of China (Grant No. 62276251), and the Joint Lab of Shenzhen Key Laboratory of Computer Vision and Pattern Recognition.

\clearpage
\newpage
{
    \small
    \bibliographystyle{ieeenat_fullname}
    \bibliography{main}
}

\clearpage
\setcounter{page}{1}
\maketitlesupplementary

\noindent In this supplementary file, we provide the following materials:

\begin{enumerate}
    \item Details of classifier.
    \item Descriptions of DilationBench and HumanBench.
    \item Comparison with lora-based few-step image inpainting models.
    \item Discussion of FID and user study.
    \item More qualitative comparisons on DilationBench and HumanBench.
    \item More visual results for ablation studies.
\end{enumerate}

\section{Details of Classifier}

\begin{figure}[t]
\centering
\small 
\begin{minipage}[t]{0.98\linewidth}
\centering
\includegraphics[width=1.0\columnwidth]{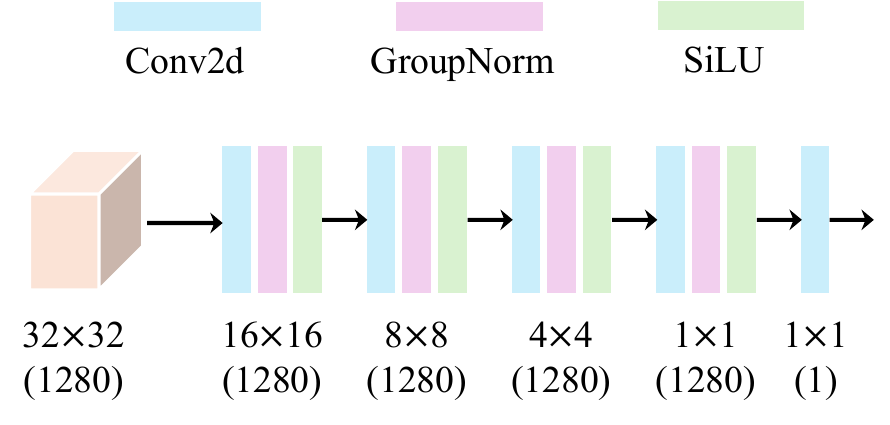}
\end{minipage}
\centering
% \vspace{-10pt}
\caption{Network structure of classifier.}
\vspace{-5pt}
\label{fig:classifier}
\end{figure}

As mentioned in the main paper, TurboFill integrates both GAN and diffusion losses to enhance image realism and style coherence. To compute the GAN loss, we map the feature output from the assistant encoder into a one-dimensional vector using a classifier, shown in 
Fig.~\ref{fig:classifier}. Specifically, the classifier consists of Conv2d layers, GroupNorm layers~\cite{wu2018group}, and SiLU activation layers~\cite{elfwing2018silu}. Starting with the feature map from the assistant encoder, the classifier uses $5$ Conv2d layers to progressively reduce the spatial dimensions from from $32 \times 32$ to $1 \times 1$.

\begin{figure}[t]
\centering
\small 
\begin{minipage}[t]{0.98\linewidth}
\centering
\includegraphics[width=1\columnwidth]{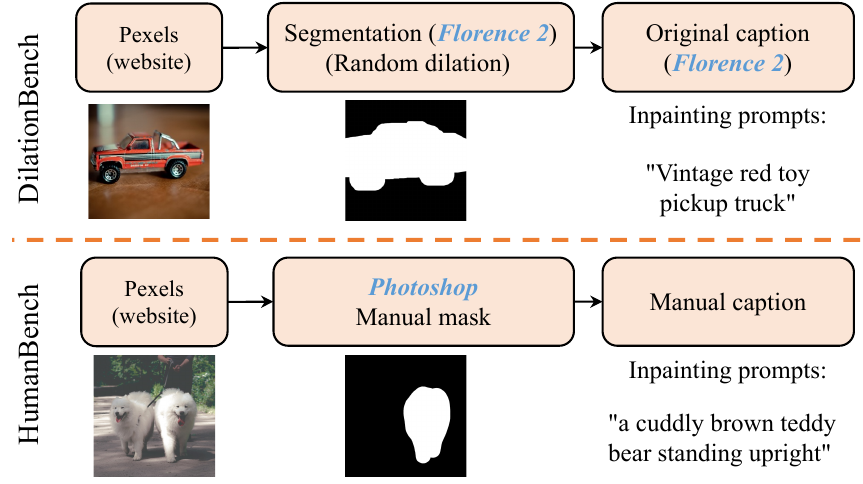}
\end{minipage}
\centering
\caption{Data collection process.}
\label{fig:dataset} 
\end{figure}

\section{Descriptions of DilationBench and HumanBench}

As shown in Fig.~\ref{fig:dataset}, the images of DilationBench and HumanBench are crawled from Pexels\textsuperscript{1} \footnote{\url{https://www.pexels.com/}}. 

For dilationBench, we employ Florence2~\cite{xiao2024florence} to perform the dense region caption task using the prompt \textless DENSE\_REGION\_CAPTION\textgreater. This task localizes primary objects in the images and generates concise textual descriptions for them. These descriptions are subsequently fed into SAM~\cite{ravi2024sam} to extract the corresponding segmentation masks. Based on the obtained masks, we first perform the erosion operation using an $8 \times 8$ kernel of ones (all-ones kernel), followed by a dilation operation to generate the final segmentation mask. Each operation is applied for 2 iterations. The inpainting prompt describes the content within the segmentation region of the original image.

For humanbench, we adopt the photoshop and manually label the segmentation mask. The inpainting prompt is also manually written.

% \usepackage{multirow}
\begin{table*}[]
\resizebox{0.98\textwidth}{!}{
\begin{tabular}{cc|ccc|ccc|c}
\hline
\multicolumn{2}{c|}{Metrics}                                                 & \multicolumn{3}{c|}{Mask Region Quality}            & \multicolumn{3}{c|}{Whole Image Quality}            & Text-Align      \\ \hline
\multicolumn{2}{c|}{Method}                                                  & Q-Align         & CLIPIQA+        & TOPIQ           & Q-Align         & CLIPIQA         & TOPIQ           & CLIP Sim        \\ \hline
\multicolumn{1}{c|}{\multirow{6}{*}{Multi-Step (50 steps)}} & BLD             & 4.1836          & 0.6874          & 5.1781          & 4.3320          & 0.6791          & 5.4662          & 24.884          \\
\multicolumn{1}{c|}{}                                      & HD-Painter      & 3.9727          & 0.6611          & 4.9345          & 4.4805          & 0.6439          & 5.4566          & 25.585          \\
\multicolumn{1}{c|}{}                                      & SDXL-Inpainting & 4.2461          & 0.6666          & 5.1533          & 4.6172          & 0.6695          & 5.5409          & 24.848          \\
\multicolumn{1}{c|}{}                                      & BrushNet-Rand   & 4.2617          & 0.6729          & 5.1738          & 4.6289          & 0.6818          & 5.5779          & 21.634          \\
\multicolumn{1}{c|}{}                                      & BrushNet* & 4.4492          & 0.7139          & 5.2744          & 4.6211          & 0.6958          & 5.5658          & 25.389          \\
\multicolumn{1}{c|}{}                                      & PowerPaint V2   & \textbf{\textcolor{red}{4.7773}} & \textbf{\textcolor{red}{0.7765}} & \textbf{\textcolor{red}{5.5468}} & \textbf{\textcolor{red}{4.7227}} & \textbf{\textcolor{red}{0.7371}} & \textbf{\textcolor{red}{5.6474}} & \textbf{\textcolor{red}{26.256}} \\ \hline
\multicolumn{1}{c|}{\multirow{6}{*}{Few-Step (4 steps)}}    & BLD             & 3.5469          & 0.5552          & 4.9605          & 4.0312          & 0.6321          & 5.4149          & 24.677          \\
\multicolumn{1}{c|}{}                                      & SDXL-Inpainting & 3.5469          & 0.5024          & 5.0080          & 4.1289          & 0.5930          & 5.4034          & 24.726          \\
\multicolumn{1}{c|}{}                                      & PowerPaint V2   & 2.7949          & 0.5958          & 4.8472          & 3.3164          & 0.6366          & 5.1755          & 22.279          \\
\multicolumn{1}{c|}{}                                      & BrushNet-Rand   & 4.1602          & 0.6538          & 4.9927          & 4.5547          & 0.6608          & 5.4599          & 21.831          \\
\multicolumn{1}{c|}{}                                      & BrushNet*       & 4.1836          & 0.6572          & 4.9939          & 4.4492          & 0.6427          & 5.3961          & 25.341          \\
\multicolumn{1}{c|}{}                                      & Ours            & \textbf{\textcolor{blue}{4.5703}} & \textbf{\textcolor{blue}{0.7332}} & \textbf{\textcolor{blue}{5.2753}} & \textbf{\textcolor{blue}{4.7188}} & \textbf{\textcolor{blue}{0.7111}} & \textbf{\textcolor{blue}{5.5392}} & \textbf{\textcolor{blue}{25.352}} \\ \hline
\end{tabular}
}
\caption{Quantitative comparisons among TurboFill and other diffusion based inpainting models in DilationBench. \textbf{\textcolor{red}{Red}} and \textbf{\textcolor{blue}{blue}} indicates the best multi-step and the best few-step performances, respectively.}
\label{tab:dilationbench}
\end{table*}

% \usepackage{multirow}
\begin{table*}[]
\resizebox{0.98\textwidth}{!}{
\begin{tabular}{cc|ccc|ccc|c}
\hline
\multicolumn{2}{c|}{Metrics}                                                 & \multicolumn{3}{c|}{Mask Region Quality}            & \multicolumn{3}{c|}{Whole Image Quality}            & Text-Align      \\ \hline
\multicolumn{2}{c|}{Method}                                                  & Q-Align         & CLIPIQA+        & TOPIQ           & Q-Align         & CLIPIQA         & TOPIQ           & CLIP Sim        \\ \hline
\multicolumn{1}{c|}{\multirow{5}{*}{Multi-Step (50 steps)}} & BLD     & 4.1523          & 0.6908          & 5.2996          & 3.9062          & 0.6579          & 5.4199          & 24.767          \\
\multicolumn{1}{c|}{}                                      & HD-Painter      & 3.9844          & 0.6493          & 5.0873          & 4.2070          & 0.6119          & 5.4381          & 25.996          \\
\multicolumn{1}{c|}{}                                      & SDXL-Inpainting & 3.9551          & 0.6405          & 5.2040          & 4.0469          & 0.6288          & 5.4565          & 24.041          \\
\multicolumn{1}{c|}{}                                      & BrushNet*       & 4.2578          & 0.7063          & 5.4340          & 4.0898          & 0.6694          & 5.5480          & 25.366          \\
\multicolumn{1}{c|}{}                                      & PowerPaint V2   & \textbf{\textcolor{red}{4.5586}} & \textbf{\textcolor{red}{0.7529}} & \textbf{\textcolor{red}{5.5475}} & \textbf{\textcolor{red}{4.3633}} & \textbf{\textcolor{red}{0.7112}} & \textbf{\textcolor{red}{5.6338}} & \textbf{\textcolor{red}{26.264}} \\ \hline
\multicolumn{1}{c|}{\multirow{5}{*}{Few-Step (4 steps)}}    & BLD             & 3.8438          & 0.5963          & 5.1622          & 3.8242          & 0.6139          & 5.3520          & 25.244         \\
\multicolumn{1}{c|}{}                                      & SDXL-Inpainting & 3.1816          & 0.4363          & 4.9627          & 3.4805          & 0.5113          & 5.2137          & 24.057         \\
\multicolumn{1}{c|}{}                                      & PowerPaint V2   & 2.7012          & 0.5641          & 4.7546          & 2.9180          & 0.5855          & 4.9632          & 20.847         \\
\multicolumn{1}{c|}{}                                      & BrushNet*       & 4.0508          & 0.6327          & 5.1041          & 4.1484          & 0.6003          & 5.3505          & 25.473          \\
\multicolumn{1}{c|}{}                                      & Ours            & \textbf{\textcolor{blue}{4.4727}} & \textbf{\textcolor{blue}{0.7257}} & \textbf{\textcolor{blue}}\textbf{\textcolor{blue}{5.3865}} & \textbf{\textcolor{blue}{4.3203}} & \textbf{\textcolor{blue}{0.6822}} & \textbf{\textcolor{blue}{5.4992}} & \textbf{\textcolor{blue}{25.710}} \\ \hline
\end{tabular}
}
\caption{Quantitative comparisons among TurboFill and other diffusion based inpainting models in HumanBench. \textbf{\textcolor{red}{Red}} and \textbf{\textcolor{blue}{blue}} indicates the best multi-step and the best few-step performances, respectively.}
\label{tab:humanbench}
\vspace{-5pt}
\end{table*}

\section{Comparison with LoRA-based Few-Step Image Inpainting Models}

For inpainting methods~\cite{avrahami2023bld, sdxl_inpaint} that train the base model, directly replacing the base model with few-step diffusion models leads to poor results. However, these models can benefit from acceleration by using few-step LoRA~\cite{hu2021lora}. Specifically, for BLD and SDXL-Inpainting, we utilize the 4-step LoRA released by DMD2~\cite{yin2024dmd2}, combined with the LCMScheduler~\cite{luo2023latent}, to construct 4-step versions of BLD~\cite{avrahami2023bld} and SDXL-Inpainting~\cite{sdxl_inpaint}. 
PowerPaint V2 is based on SD 1.5~\cite{rombach2022sd} and BrushNet. Since DMD2 only offers a LoRA version compatible with SDXL, we use the acceleration LoRA provided by HyperSD~\cite{ren2024hyper}. The quantitative results for these three LoRA-based few-step inpainting models are presented in the Tab.~\ref{tab:dilationbench} and Tab.~\ref{tab:humanbench}.

The results on DilationBench and HumanBench reveal that the 4-step models accelerated using LoRA exhibit a significant performance drop compared to the original 50-step models, with the 4-step PowerPaint V2 performing the worst. Moreover, the acceleration achieved with LoRA falls far short of that achieved by models capable of replacing the base model, such as BrushNet. Among these few-step image inpainting models, TurboFill demonstrates the best performance.

\begin{figure*}[t]
\centering
\small 
\begin{minipage}[t]{0.98\linewidth}
\centering
\includegraphics[width=1.0\columnwidth]{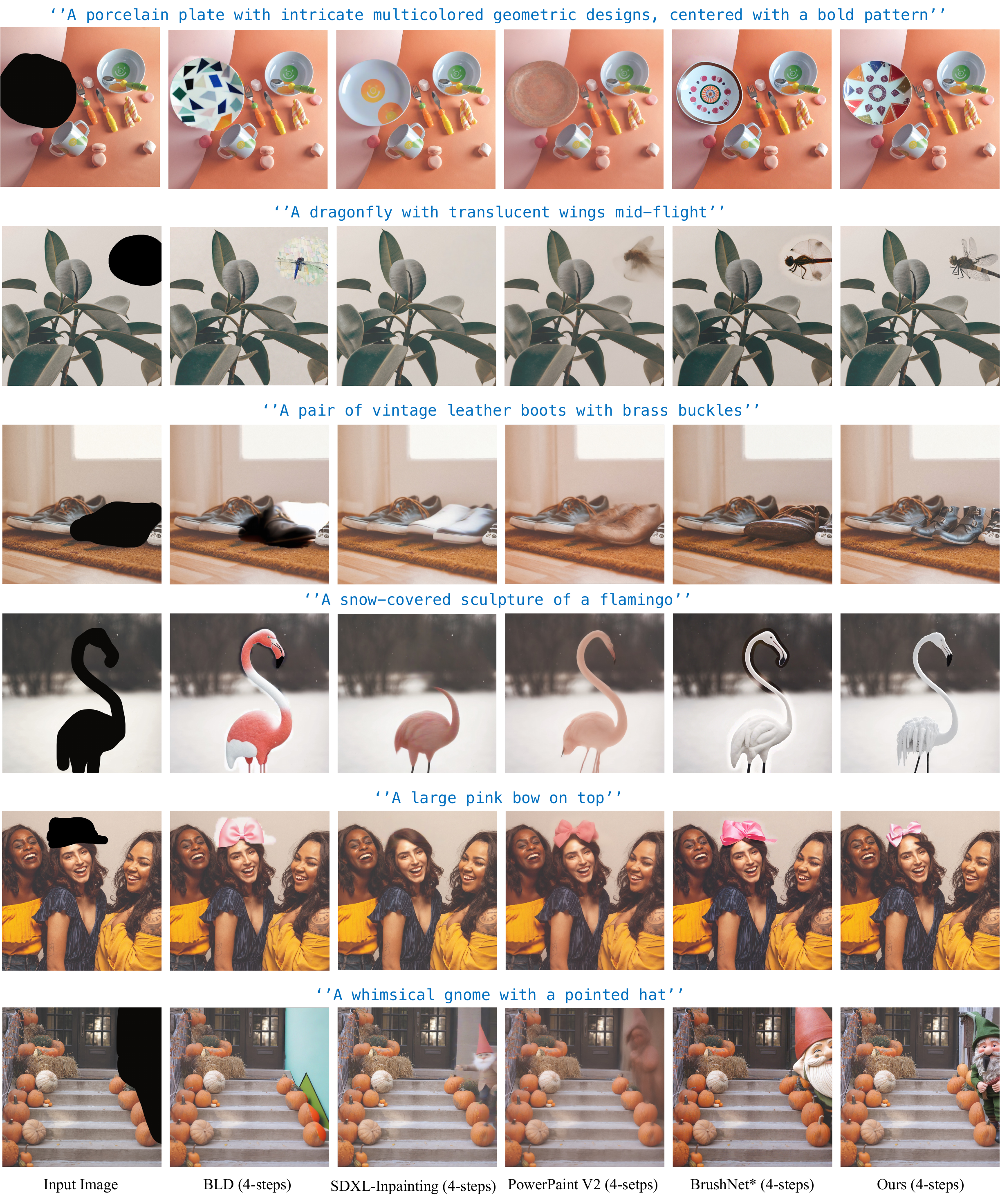}
\end{minipage}
\centering
% \vspace{-10pt}
\caption{Comparison of few-step image inpainting methods on DilationBench. Compared to other few-step image inpainting models, TurboFill produces results that align more effectively with the prompt. Furthermore, TurboFill generates more realistic details and textures while achieving effective scene harmonization. (\textbf{Zoom in for best view})}
% \vspace{10pt}
\label{fig:few_step_comparison}
\end{figure*}

We also visualize the comparison of few-step image inpainting methods in Fig.~\ref{fig:few_step_comparison}. It is evident that SDXL-Inpainting and PowerPaint V2 produce results with poor details and often fail to align with the prompt (e.g., rows 2 and 4). The results of BLD are slightly better, but they still exhibit noticeable artifacts (e.g., rows 2, 3 and 5) and occasionally generate outputs completely misaligned with the prompt (e.g., row 6). Similarly, BrushNet* sometimes aligns only partially with the prompt (e.g., rows 1 and 3). In contrast, TurboFill consistently produces prompt-aligned results with realistic details, rich textures, and seamless scene harmonization.

\begin{table*}[]
\centering
\resizebox{0.60\textwidth}{!}{
\begin{tabular}{c|cc|cc}
\hline
Methods          & \multicolumn{2}{c|}{Multi-step ($50$ steps)} & \multicolumn{2}{c}{Few-step ($4$ steps)} \\ \hline
Metrics          & BrushNet*      & PowerPaint V2    & BrushNet*     & TurboFill    \\ \hline
FID $\downarrow$ (5 images)   & \textbf{29.3350}        & 37.4552        & 31.8095       & 33.1541      \\
FID $\downarrow$ (300 images) & 90.6107        & 91.9412        & 89.8712       & \textbf{89.8300 }     \\ \hline
\end{tabular}
}
\caption{Evaluation results based on the FID metric. The FID scores exhibit significant variability when applied to different GT datasets, indicating that FID is not a suitable metric for assessing diffusion-based image inpainting tasks.}
\label{tab:fid}
\vspace{-5pt}
\end{table*}

\begin{figure*}[t]
\centering
\small 
\begin{minipage}[t]{0.95\linewidth}
\centering
\includegraphics[width=1.0\columnwidth]{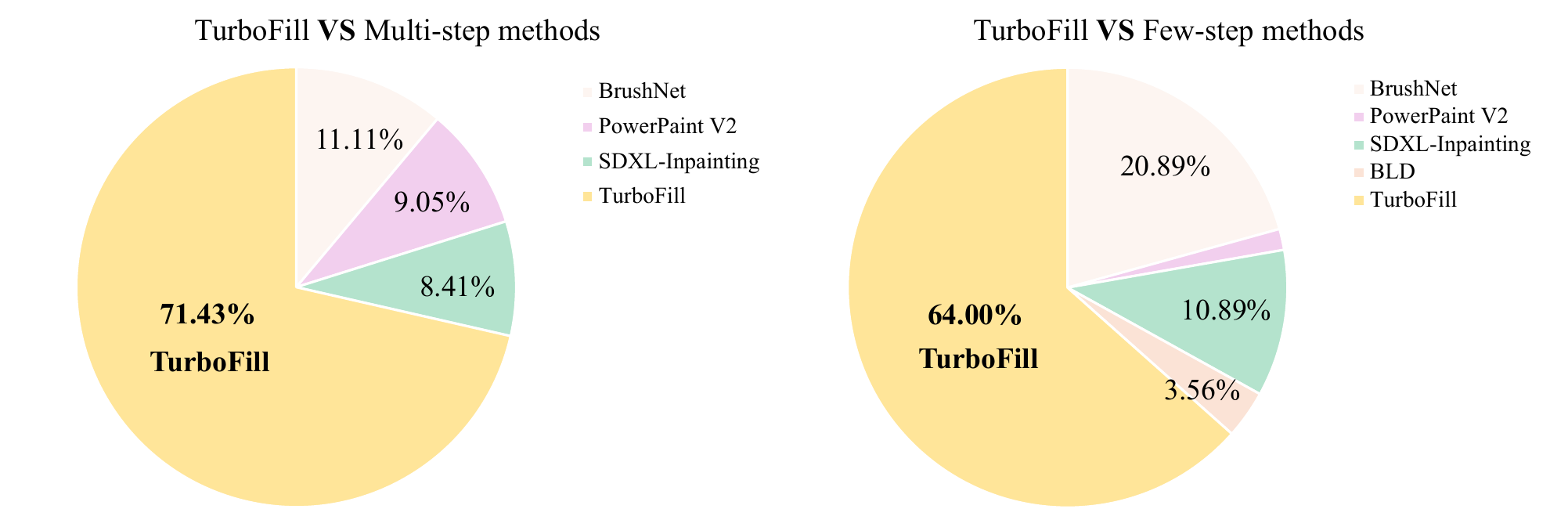}
\end{minipage}
\centering
% \vspace{-10pt}
\caption{The results of user studies. We design two separate user studies: one comparing TurboFill with multi-step methods (left pie chart) and the other comparing TurboFill with few-step methods (right pie chart). It is evident that TurboFill's results are more favored by participants.}
\vspace{-6pt}
\label{fig:user_study}
\end{figure*}

\section{Discussion of FID and User Study}

Fréchet Inception Distance (FID), which measures the distance between feature distributions of generated images and a ground truth (GT) dataset, is adopted as a primary evaluation metric in many Image Inpainting works~\cite{chen2025catdiffusion, chen2024powerpaint}. However, we find that FID does not reliably reflect the visual quality of results.

To investigate this issue, based on $5$ images (Figure 3 in the main paper), we calculate FID scores of four methods. The results as shown in Tab.~\ref{tab:fid}. Our analysis shows that when we use the original images ($5$ images) as GT, BrushNet* ($50$ steps) achieves the best performance while TurboFill is visually better. However, when we switch to a different GT dataset ($300$ images within DilationBench), TurboFill performs the best. This indicates that FID is highly sensitive to the choice of GT and, therefore, is not a reliable metric for evaluating inpainting results.

Considering that the ultimate goal of existing metrics is to align with human rater preferences, we directly conduct the user study to evaluate the results of different image inpainting methods. Specifically, we design two separate user studies: one comparing TurboFill with multi-step methods and the other comparing TurboFill with few-step methods. For each comparison group in the user study, we randomly shuffle the order of results from all methods and ask participants to select the highest-quality and most natural image, aligned with the prompt. Each user study includes $30$ groups of images, and we invite $20$ participants to take part in the evaluation.

The results of user study are shown in Fig.~\ref{fig:user_study}. When comparing against multi-step image inpainting methods, including BrushNet~\cite{ju2024brushnet}, PowerPaint V2~\cite{chen2024powerpaint}, and SDXL-Inpainting~\cite{sdxl_inpaint}, over $70\%$ of participants prefer TurboFill, as shown in the left pie chart. They highlight its ability to produce more natural and detail-rich results. When comparing against few-step methods, including BrushNet, PowerPaint V2, SDXL-Inpainting, and BLD~\cite{avrahami2023bld}, TurboFill is favored by approximately $64\%$ of participants, as shown in the right pie chart. Notably, PowerPaint V2, when accelerated with HyperSD's LoRA~\cite{ren2024hyper}, generates blurred results with a lack of high-frequency details, as seen in Fig.~\ref{fig:few_step_comparison}, which likely contributes to its lower preference.

\section{More Qualitative Comparisons}

\begin{figure*}[t]
\centering
\small 
\begin{minipage}[t]{0.95\linewidth}
\centering
\includegraphics[width=1.0\columnwidth]{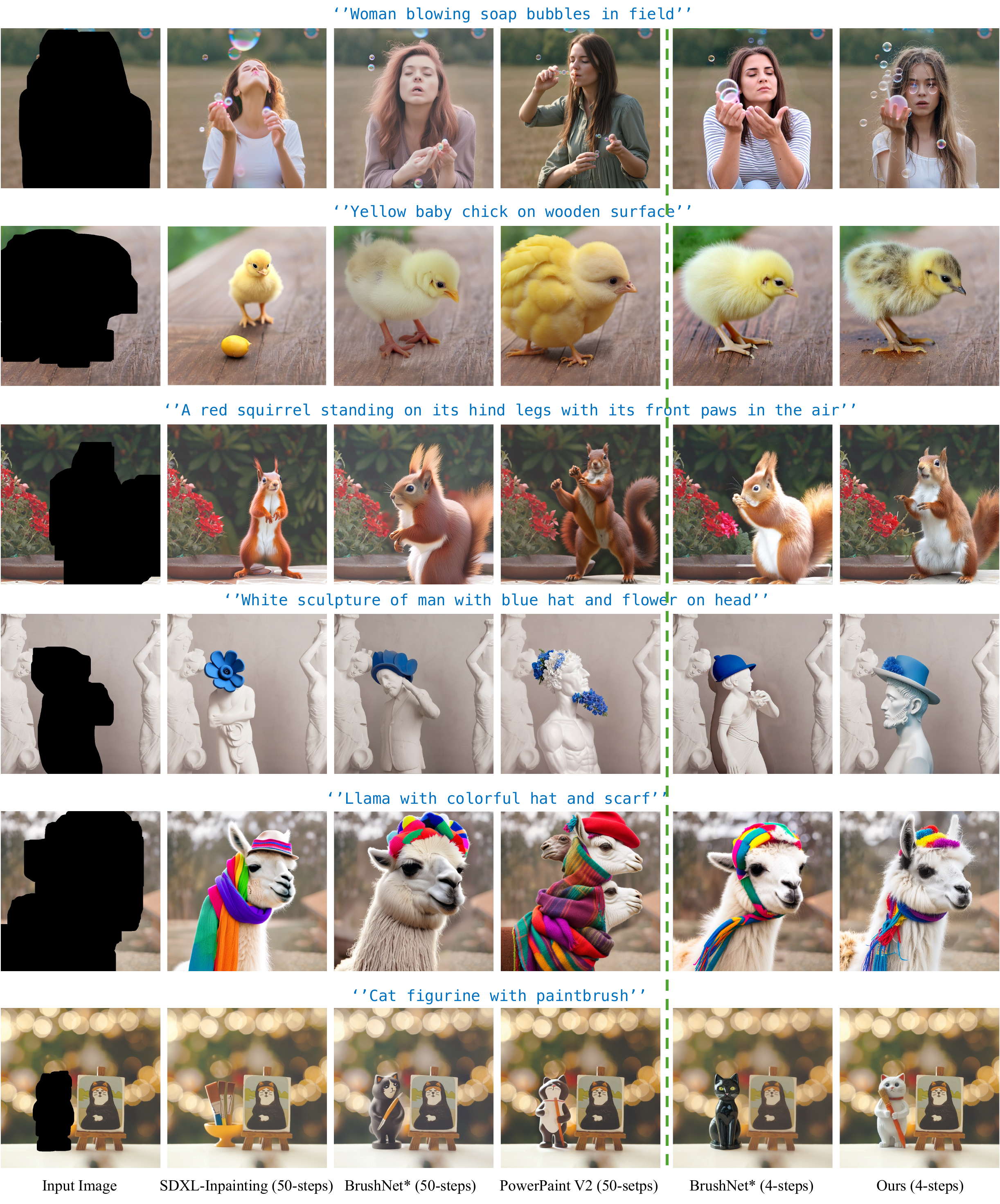}
\end{minipage}
\centering
% \vspace{-10pt}
\caption{Comparison of previous inpainting methods and BrushNet on DilationBench. Compared to other methods, TurboFill generates more realistic details and textures in just 4 steps, while achieving good scene harmonization. (\textbf{Zoom in for best view})}
% \vspace{10pt}
\label{fig:dilation_comparison1}
\end{figure*}

\begin{figure*}[t]
\centering
\small 
\begin{minipage}[t]{0.95\linewidth}
\centering
\includegraphics[width=1.0\columnwidth]{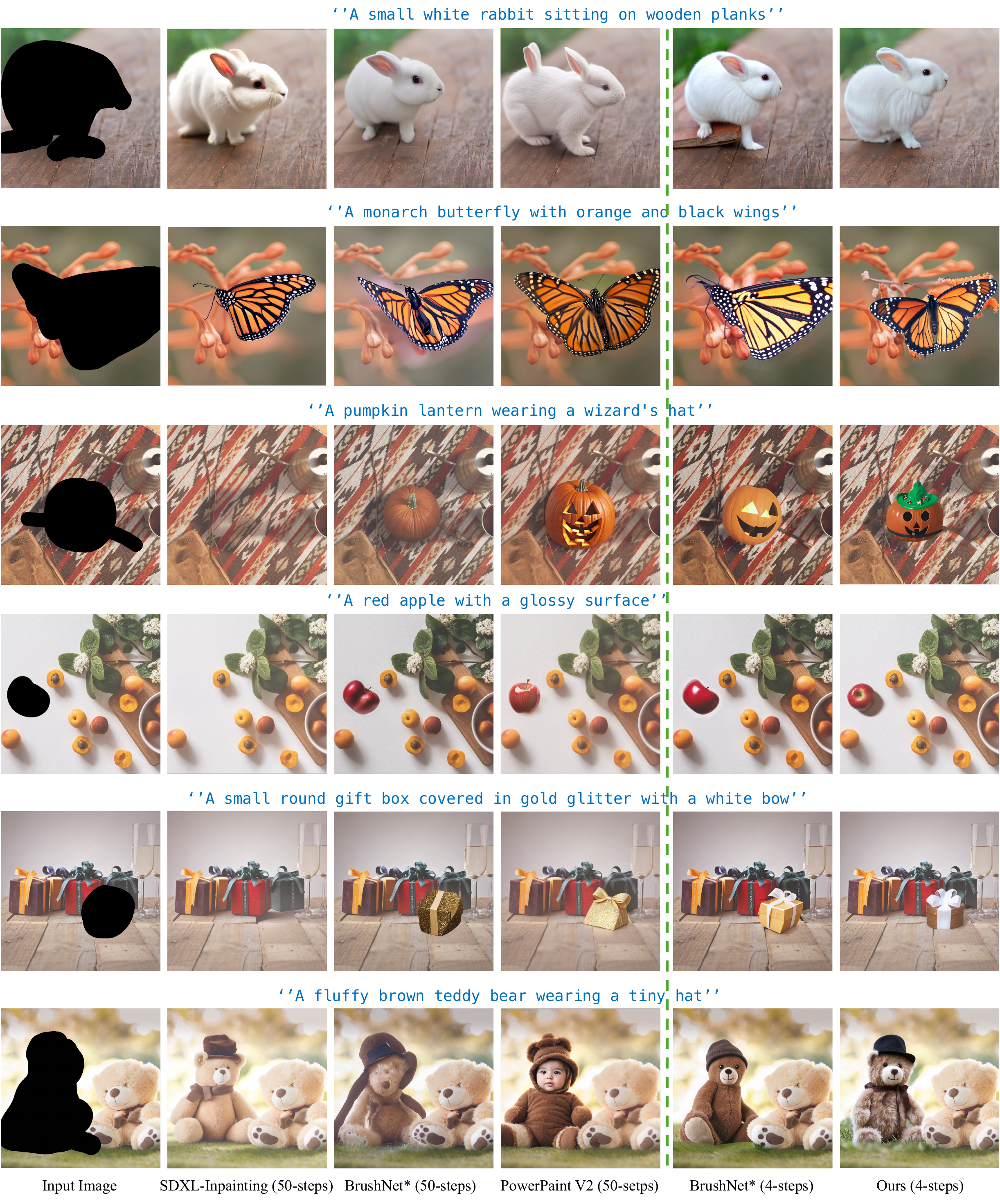}
\end{minipage}
\centering
% \vspace{-10pt}
\caption{Comparison of previous inpainting methods and BrushNet on HumanBench. Compared to other methods, TurboFill generates more realistic details and textures in just 4 steps, while achieving good scene harmonization. (\textbf{Zoom in for best view})}
% \vspace{10pt}
\label{fig:human_comparison1}
\end{figure*}

\begin{figure*}[t]
\centering
\small 
\begin{minipage}[t]{0.95\linewidth}
\centering
\includegraphics[width=1.0\columnwidth]{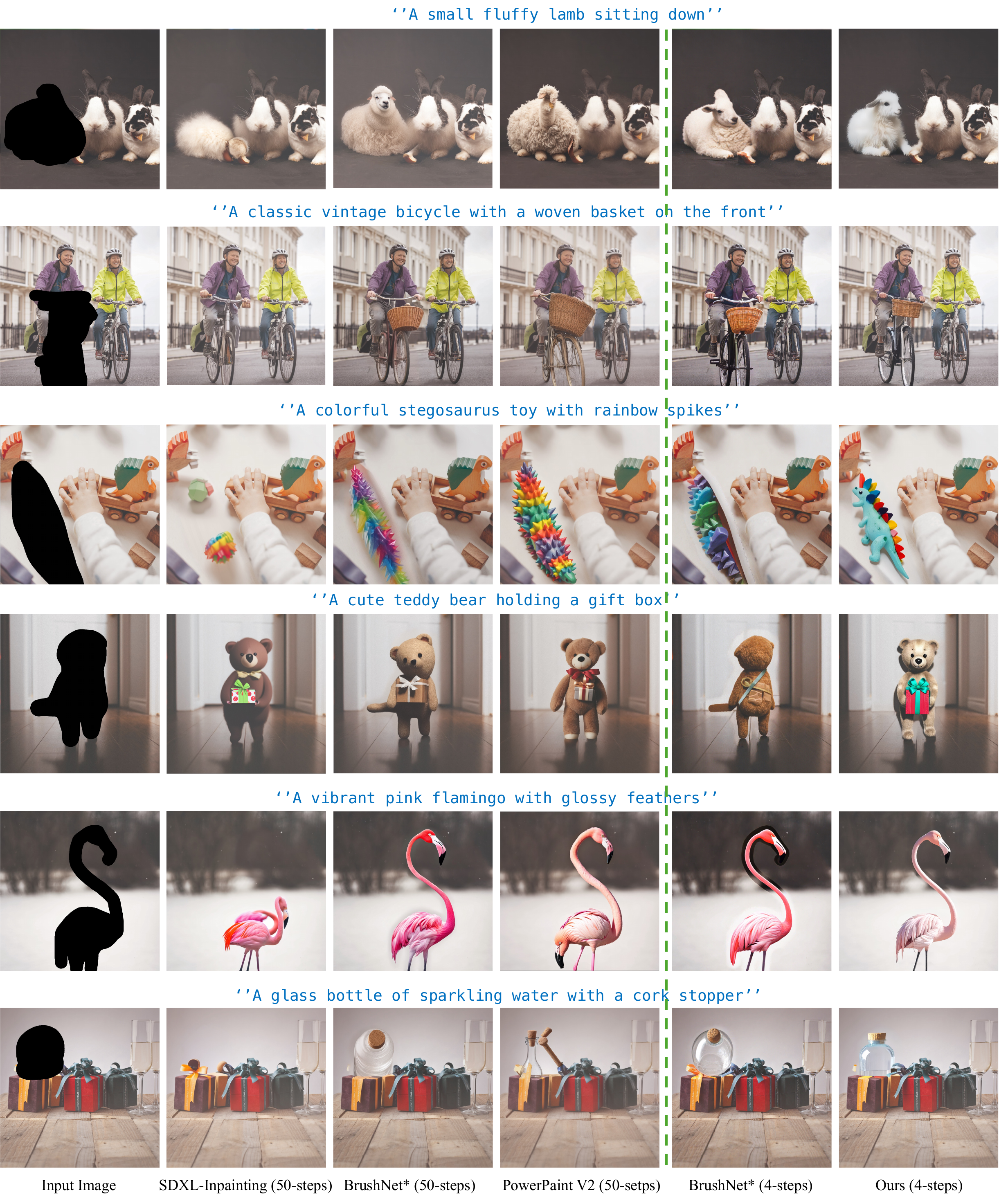}
\end{minipage}
\centering
% \vspace{-10pt}
\caption{Comparison of previous inpainting methods and BrushNet on HumanBench. Compared to other methods, TurboFill generates more realistic details and textures in just 4 steps, while achieving good scene harmonization. (\textbf{Zoom in for best view})}
% \vspace{10pt}
\label{fig:human_comparison2}
\end{figure*}

We present additional qualitative comparisons based on DilationBench (Fig.~\ref{fig:dilation_comparison1}) and HumanBench (Fig.~\ref{fig:human_comparison1}, Fig.~\ref{fig:human_comparison2}). For DilationBench, SDXL-Inpainting and BrushNet* ($50$ steps) often only partially reflect the prompt content in their results (e.g., rows 4 and 6). PowerPaint V2 exhibits significant distortion issues (e.g., rows 1, 3, and 5), while BrushNet* (4 steps) occasionally produces oversaturated results (e.g., row 3). In contrast, our method demonstrates excellent detail preservation (e.g., the fur of animals) and achieves a harmonious overall image without overexposure.

For DilationBench, as shown in Fig.~\ref{fig:human_comparison1} and Fig.~\ref{fig:human_comparison2}, it is observed that SDXL-Inpainting often fills the background into the inpainted areas (e.g., rows 3, 4, and 5 (Fig.~\ref{fig:human_comparison1})). In comparison, BrushNet* (50 steps) frequently produces results misaligned with the prompt (e.g., rows 3 and 5 (Fig.~\ref{fig:human_comparison1}), row 3 (Fig.~\ref{fig:human_comparison2})) and introduces noticeable artifacts (e.g., row 2 (Fig.~\ref{fig:human_comparison1})). PowerPaint V2, on the other hand, generates results with significant distortions (e.g., rows 1 and 3 (Fig.~\ref{fig:human_comparison1}), row 5 (Fig.~\ref{fig:human_comparison2})). BrushNet* (4 steps) exhibits evident overexposure issues (e.g., rows 1, 2, and 4 (Fig.~\ref{fig:human_comparison1}), row 5 (Fig.~\ref{fig:human_comparison2})). Unlike other methods, TurboFill produces results that are more harmonious, with richer details, and effectively adheres to the prompt.

\section{Visual Results for Ablation Studies}

Starting from TurboFill, we remove $\mathcal{L}_{\mathrm{BG}}$, $\mathcal{L}^{F}_{\mathrm{Diff}}$, and $\mathcal{L}^{R}_{\mathrm{Diff}}$ in sequence, with qualitative results shown in Fig.~\ref{fig:ablation}. In the visualizations, we see that without $\mathcal{L}_{\mathrm{BG}}$, the color of background region changes noticeably, creating a sharp boundary between the fill-in and background regions. Further removing $\mathcal{L}^{F}_{\mathrm{Diff}}$ introduces conflicting elements (i.e., house) in the fill-in region, suggesting the discriminator fails to fully capture the holistic scene. Finally, without $\mathcal{L}^{R}_{\mathrm{Diff}}$, relying only on GAN loss, the inpainted images exhibit not only inconsistencies with the background but also poor texture and detail. This highlights that GAN loss alone struggles to close the gap between fake and real latents. Only when combining $\mathcal{L}^{F}_{\mathrm{Diff}}$, $\mathcal{L}^{R}_{\mathrm{Diff}}$, and $\mathcal{L}_{\mathrm{BG}}$ in GAN training does the model achieve enhanced texture, detail, and effective scene harmonization between fill-in and background regions.

\begin{figure*}[t]
\centering
\small 
\begin{minipage}[t]{0.98\linewidth}
\centering
\includegraphics[width=1.0\columnwidth]{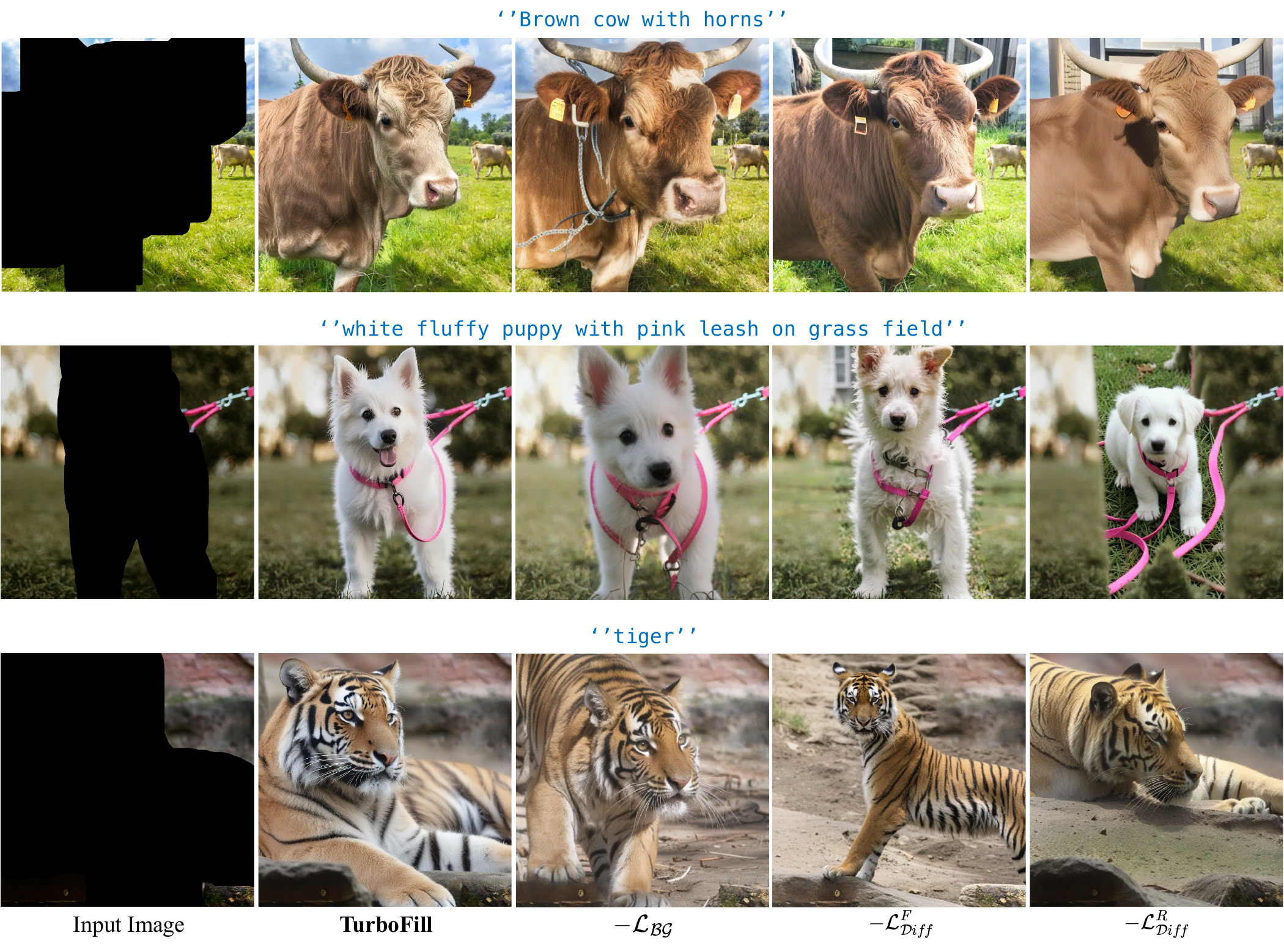}
\end{minipage}
\centering
% \vspace{-10pt}
\caption{The effectiveness of different losses. From left to right, we progressively remove specific losses. (\textbf{Zoom in for best view})}
% \vspace{10pt}
\label{fig:ablation}
\end{figure*}


\end{document}